\documentclass[11pt]{article}

\usepackage[final]{acl}

\usepackage{times}
\usepackage{latexsym}
\usepackage[T1]{fontenc}
\usepackage[utf8]{inputenc}
\usepackage{microtype}
\usepackage{inconsolata}
\usepackage{marvosym}

\usepackage{graphicx}
\usepackage{url}
\usepackage{xurl}
\usepackage{booktabs}
\usepackage{amsfonts}
\usepackage{nicefrac}
\usepackage{xcolor}
\usepackage{multirow}
\usepackage{subcaption}
\usepackage[most]{tcolorbox}
\usepackage{makecell}
\usepackage{amsmath,amssymb}
\usepackage{array}
\usepackage{enumitem}
\usepackage{fvextra}
\usepackage{float}
\usepackage{needspace}
\usepackage{longtable}
\usepackage{titlesec}
\newcolumntype{L}[1]{>{\raggedright\arraybackslash}p{#1}}
\newcommand{\blue}[1]{\textcolor{blue}{#1}}
\newcommand{\red}[1]{\textcolor{red}{#1}}

\definecolor{PromptBlue}{RGB}{70,130,180} 

\newtcolorbox{promptbox}[1]{
  enhanced,
  breakable,
  colback=gray!3,
  colframe=PromptBlue,
  boxrule=0.4pt,
  arc=1mm,
  left=1.5mm,
  right=1.5mm,
  top=1mm,
  bottom=1mm,
  title={#1},
  fonttitle=\bfseries,
  fontupper=\small,
  before skip=6pt,
  after skip=6pt
}

\DefineVerbatimEnvironment{PromptCode}{Verbatim}{
  breaklines=true,
  breakanywhere=false,
  commandchars=\\\^\~,
  breaksymbolleft={},
  breaksymbolright={},
  fontsize=\footnotesize,
  baselinestretch=0.9
}

\newcommand{\code}[1]{\texttt{\detokenize{#1}}}
\definecolor{PromptVarBlue}{RGB}{0,82,155}
\newcommand{\pvar}[1]{\textcolor{PromptVarBlue}{\texttt{\{\detokenize{#1}\}}}}

\newcommand{\TACE}{TACE}

\title{\textsc{GameCommBench}: A Unified Benchmark and Type-Aware Evaluation for AI-Generated Game Commentary}

\author{
  \textbf{Qirui Zheng\textsuperscript{1}}, 
  \textbf{Zhengteng Lin\textsuperscript{2}}, 
  \textbf{Yunyi Xiao\textsuperscript{2}}, 
  \textbf{Junhao Li\textsuperscript{2}}, 
  \textbf{Keyuan Cheng\textsuperscript{1}},
  \textbf{Xingbo Wang\textsuperscript{1}}, \\
  \textbf{Yongyi Wang\textsuperscript{1}}, 
  \textbf{Lingfeng Li\textsuperscript{1}}, 
  \textbf{Yunlong Lu\textsuperscript{1}}, 
  \textbf{Wenxin Li\textsuperscript{1, \Letter}} 
  \\
  \textsuperscript{1}Peking University \hspace*{0.5cm}
  \textsuperscript{2}South China University of Technology
  \\
  \small{
    \href{mailto:zzzqr@stu.pku.edu.cn}{zzzqr@stu.pku.edu.cn} \hspace*{0.5cm} \href{mailto:lwx@pku.edu.cn}{lwx@pku.edu.cn}
  }
}

\begin{document}

\maketitle

\begin{abstract}
Game commentary is an open-ended generation task requiring multimodal perception, strategic reasoning, and contextual knowledge. Existing AI-Generated Game Commentary (AI-GGC) studies remain fragmented across games, modalities, and evaluation protocols, while overlap-based or holistic evaluators fail to capture the functional heterogeneity of commentary. We introduce GameCommBench, a unified benchmark spanning board games, sports, and esports, with commentary aligned to heterogeneous game contexts and annotated by commentary type. We further propose Type-Aware Commentary Evaluation (TACE), a structured framework for evaluating different types of commentary. We then validate TACE for reliability and human agreement, and use it to benchmark representative AI commentators. Results reveal non-uniform capability profiles, with live observation and strategic analysis emerging as major bottlenecks. Together, GameCommBench and TACE provide a diagnostic foundation for comparable and interpretable AI-GGC evaluation.
\end{abstract}

\begingroup
\renewcommand\thefootnote{}
\footnotetext{
\vspace{-1.5ex}
\href{https://anonymous.4open.science/r/GameCommBench-0382/}{\textcolor{blue}{Code}} and \href{https://huggingface.co/datasets/anonymousForBlind/GameCommBench}{\textcolor{blue}{Data}} are available here.
}
\addtocounter{footnote}{-1}
\endgroup

\section{Introduction}


Game commentary helps audiences make sense of gameplay in real time by narrating key events, interpreting important decisions, and situating moments in broader context~\citep{behrens2022you}. AI-Generated Game Commentary (AI-GGC) has therefore become an increasingly active research area with clear practical value. Beyond utility, AI-GGC is a challenging open-ended NLG task: high-quality commentary requires live observation of unfolding events, strategic analysis of decisions and consequences, and historical recall of relevant game context~\citep{zheng2025multimodal}.


Despite growing interest, existing AI-GGC resources remain fragmented across individual games, including soccer~\citep{giancola2018soccernet, deliege2021soccernet}, basketball~\citep{yu2018fine, yan2019fine}, chess~\citep{jhamtani2018learning, lee2022improving}, Go~\citep{tomlin2022understanding}, and League of Legends(LoL)~\citep{zhang2024game, wang2025commentary}. These resources have supported progress within their domains, but they remain tied to specific games, input modalities, data formats, and comment focuses. As a result, the field lacks a unified benchmark that organizes heterogeneous game contexts under a shared interface, making systematic cross-setting comparison difficult.

\begin{figure*}[t]
    \centering
    \vspace{-1.1cm}
    \includegraphics[width=\textwidth]{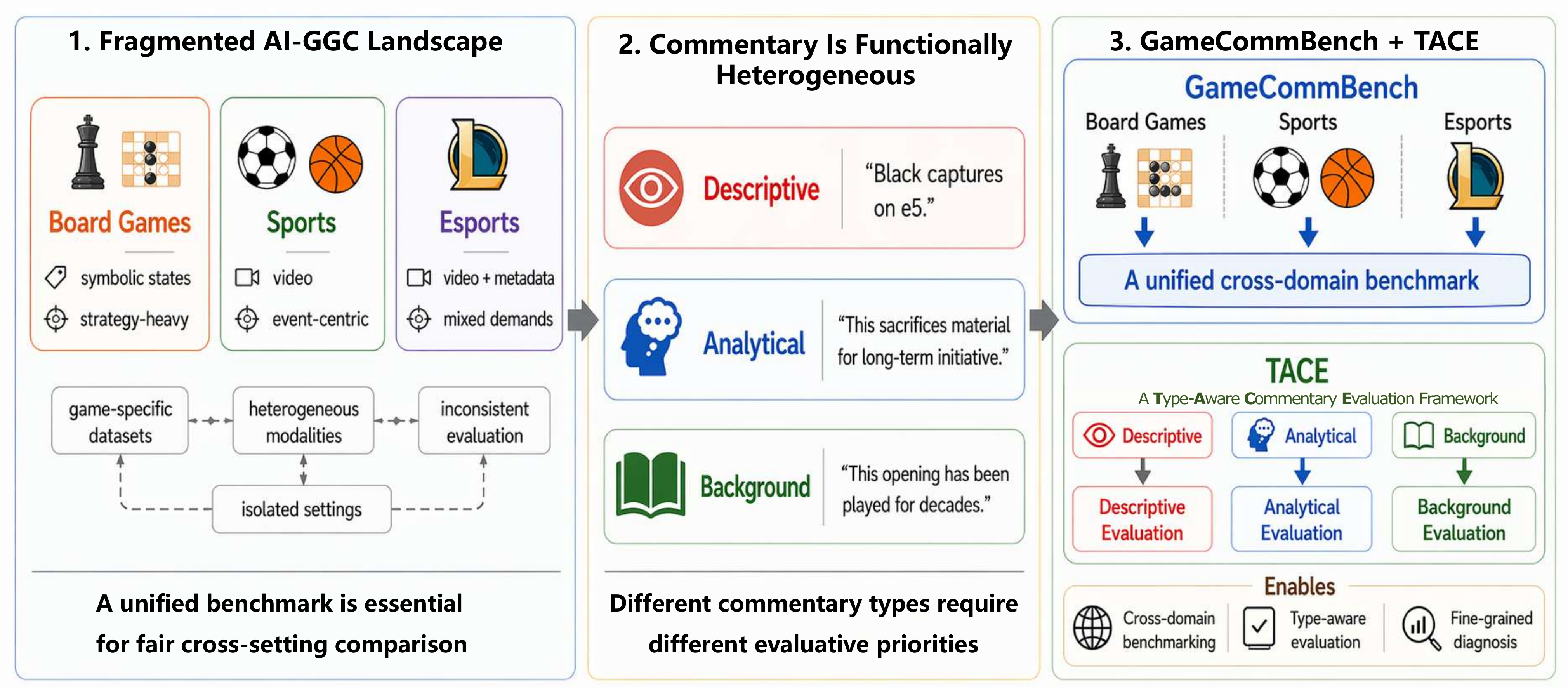}
    \vspace{-3.5ex}
    \caption{Existing AI-GGC research is fragmented across domains and modalities, and lacks effective evaluation schemes. We address these issues with \textsc{GameCommBench}, a unified cross-domain benchmark, and TACE, a commentary evaluation framework designed for descriptive, analytical, and background commentary.}
    \label{fig:teasor}
    \vspace{-1.5ex}
\end{figure*}


\vspace{-0.1ex}
Evaluating game commentary poses an even deeper challenge. Many AI-GGC studies still rely on n-gram overlap metrics~\citep{zheng2025multimodal}, such as BLEU, ROUGE, or METEOR, despite the open-ended nature of commentary: for same game state, different commentators may produce distinct yet valid outputs. Such metrics therefore penalize variation rather than measuring commentary quality. Recent LLM-based evaluators move beyond lexical overlap~\citep{liu2023g}, but existing approaches still score commentary holistically~\citep{kim2025bridging}. This is insufficient because commentary serves distinct functions, including describing game states, interpreting strategic intent, and providing broader historical context~\citep{zheng2025multimodal}. Each function requires different criteria for judging quality, so evaluation should target these functions separately rather than collapse them into a single score. AI-GGC further complicates evaluation because a given game state admits neither a unique correct commentary nor a single correct reasoning process, especially for the analytical part.

\vspace{-0.1ex}
To address these gaps, we present \textsc{GameCommBench} and Type-Aware Commentary Evaluation (TACE). \textsc{GameCommBench} brings board games, sports, and esports into a unified benchmark, while TACE evaluates different functional types of commentary with type-specific evaluation metrics. This design enables systematic comparison of AI commentators across games, and supports finer-grained diagnosis of model capabilities.

Our main contributions are as follows: (1) We construct \textsc{GameCommBench}, a unified AI-GGC benchmark with heterogeneous game contexts and type-annotated commentary. (2) We propose TACE that assesses descriptive, analytical, and background commentary with dedicated criteria instead of a single holistic score. (3) We validate TACE through repeated-run reliability analysis, human validation of LLM-mediated components, and cross-judge consistency checks. (4) We benchmark representative AI commentators and reveal non-uniform capability profiles, with live observation and strategic analysis remaining key bottlenecks.





\begin{figure*}[t]
    \centering
    \vspace{-1.2cm}
    \includegraphics[width=\linewidth]{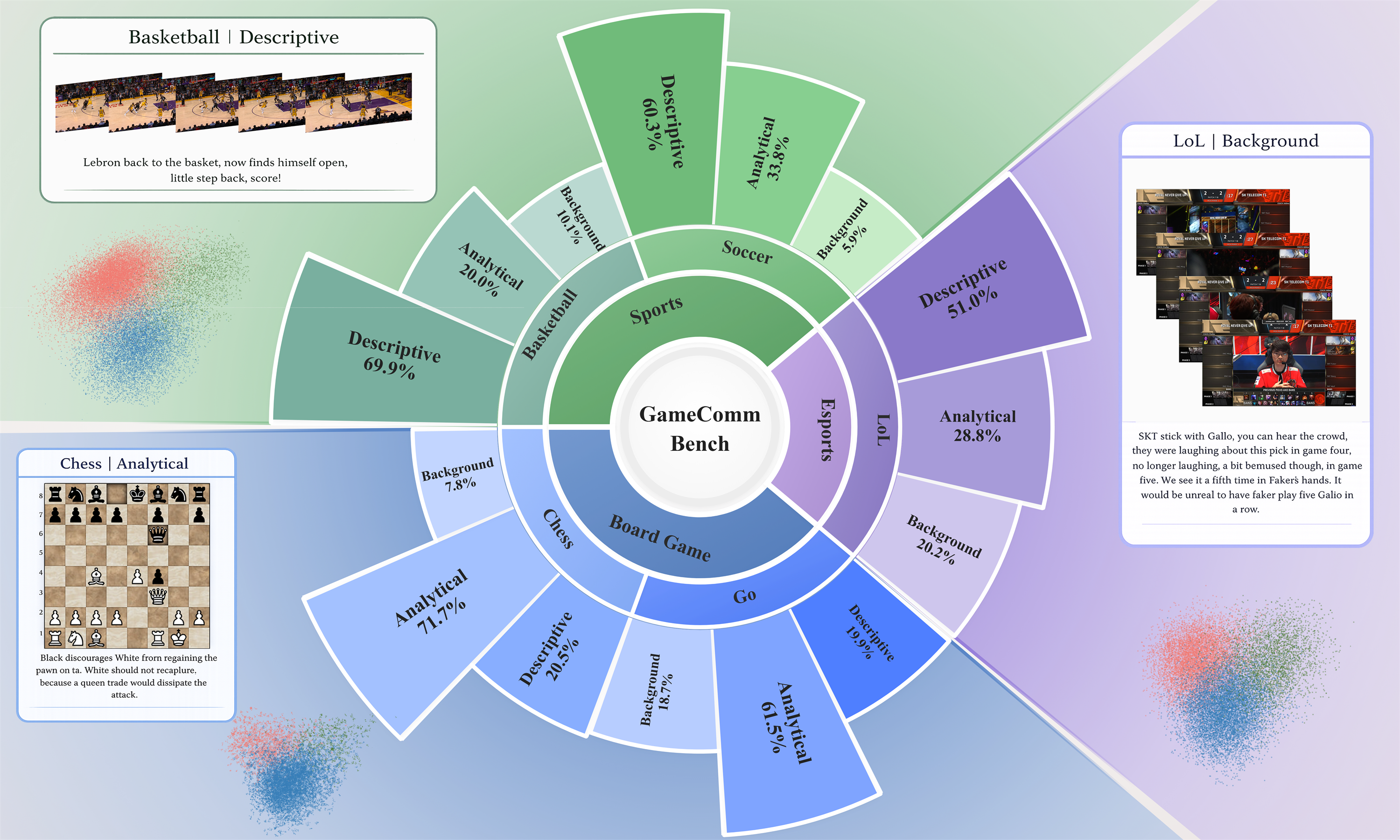}
    \vspace{-3.5ex}
    \caption{\textbf{Overview of \textsc{GameCommBench}.}
The figure summarizes benchmark coverage, representative state-commentary examples, and commentary-type distributions across games. The central wheel reports the proportions of descriptive, analytical, and background commentary for each game. Scatter plots visualize classified human commentary embeddings, with red, blue, and green denoting descriptive, analytical, and background commentary.}
    \label{overview}
    \vspace{-1ex}
\end{figure*}

\section{Benchmark Construction}

We construct \textsc{GameCommBench} as a unified AI-GGC benchmark spanning board games, sports, and esports. Figure~\ref{overview} summarizes its detailed game coverage and data composition.

\subsection{Benchmark Scope and Domain Coverage}

\textsc{GameCommBench} covers five representative games across three domains: chess and Go for board games, soccer and basketball for sports, and LoL for esports. These games are selected not simply to broaden task coverage, but to expose variation in game-state representation, game dynamics, and demands on commentator abilities.

Board games provide relatively structured symbolic states but place strong demands on long-horizon strategic analysis~\citep{lee2022improving}. Commentary in board games typically emphasizes evaluating move quality, analysing strategy, and predicting possible future moves. Sports games, by contrast, are grounded in dynamic visual scenes, requiring commentators to identify salient events, track fast-changing actions, and produce coherent narration under temporal constraints~\citep{you2025timesoccer}. Esports introduce an additional layer of complexity, combining visually crowded, fast-changing gameplay with rich API-based information. This places higher demands on commentators to integrate complex multimodal inputs for game understanding and commentary generation~\citep{zheng2025multimodal}.

\subsection{Data Collection and Processing Pipeline}

\textbf{Data Collection and Alignment.}
We collect aligned game contexts and commentary for the five games in \textsc{GameCommBench} from domain-specific sources. For chess and Go, we obtain game records and commentary from \href{https://gameknot.com/}{GameKnot} and \href{https://foxwq.com/}{Foxwq}, where comments are naturally aligned with turns. For basketball and soccer, we use broadcast videos and action labels from high-quality datasets, NSVA~\citep{wu2022sports} for basketball and SoccerReplay-1988~\citep{rao2025towards} for soccer, and transcribe spoken commentary with WhisperX~\citep{bain2023whisperx}. For LoL, we collect official video clips paired with timestamped textual commentary; the timestamps are used to retrieve fine-grained API snapshots, yielding player-level and team-level game-state information. We also collect player and team background information from the same channels as supplementary context for commentary generation.

\textbf{Data Cleaning and Quality Control.}
We combine manual screening and filtering with alignment-level repair. At the match level, we prioritize games with dense commentary, competitive importance, and broad representativeness, while maintaining team and player diversity to avoid over-concentration on recurring entities. At the alignment level, board-game commentary naturally maps to discrete game states, whereas sports and esports rely on timestamp-based alignment and are more susceptible to sentence-boundary errors. To address this, we apply an LLM-based post-processing step that merges an incomplete first utterance in a segment with the previous segment. We further remove only content that is completely unrelated to the game itself, such as off-topic chatter or broadcast-side digressions. The original raw commentary is retained in each benchmark record for traceability and future research use.

\textbf{Semantic Annotation.}
Since game commentary often mixes multiple functions, we decompose each commentary into finer semantic units and label them as descriptive, analytical, or background. This step is performed by LLM, and its agreement with human annotation is reported in Table~\ref{tab:type_decomposition_acc}. This unit-level labeling avoids forcing an entire utterance into a single type and allows TACE to evaluate each functional part separately.

\textbf{Unified Packaging.}
Finally, all retained instances are packaged into a unified record schema. Each record contains raw and processed commentary aligned with game state, type-labeled semantic units, and game-related information. This schema keeps game-specific inputs such as symbolic states, videos, and API snapshots under a shared interface for generation and comparison.

\subsection{Data Composition and Distribution}

Figure~\ref{overview} summarizes the benchmark composition across games. Since alignment units differ across domains, we report commentary-bearing turns for board games and aligned video clips for sports and esports. Specifically, the chess and Go subsets contain 2,159 and 2,857 turns, with 3,405 and 5,811 commentary sentences, respectively. In the sports domain, the soccer and basketball subsets contain 2,681 and 2,981 aligned video clips, paired with 5,893 and 6,120 sentences, respectively. For esports, the LoL subset provides 7,675 aligned clips and 10,001 sentences. Detailed commentary-type proportions for each game are shown in Figure~\ref{overview}.

Overall, \textsc{GameCommBench} exhibits substantial cross-game variation in scale, modality, and commentary composition. Board game commentary is more concentrated in analysing, sports commentary is predominantly descriptive, and esports commentary shows a more balanced distribution.

\section{Type-Aware Commentary Evaluation}

\begin{figure*}[t]
    \centering
    \vspace{-1.2cm}
    \includegraphics[width=\linewidth]{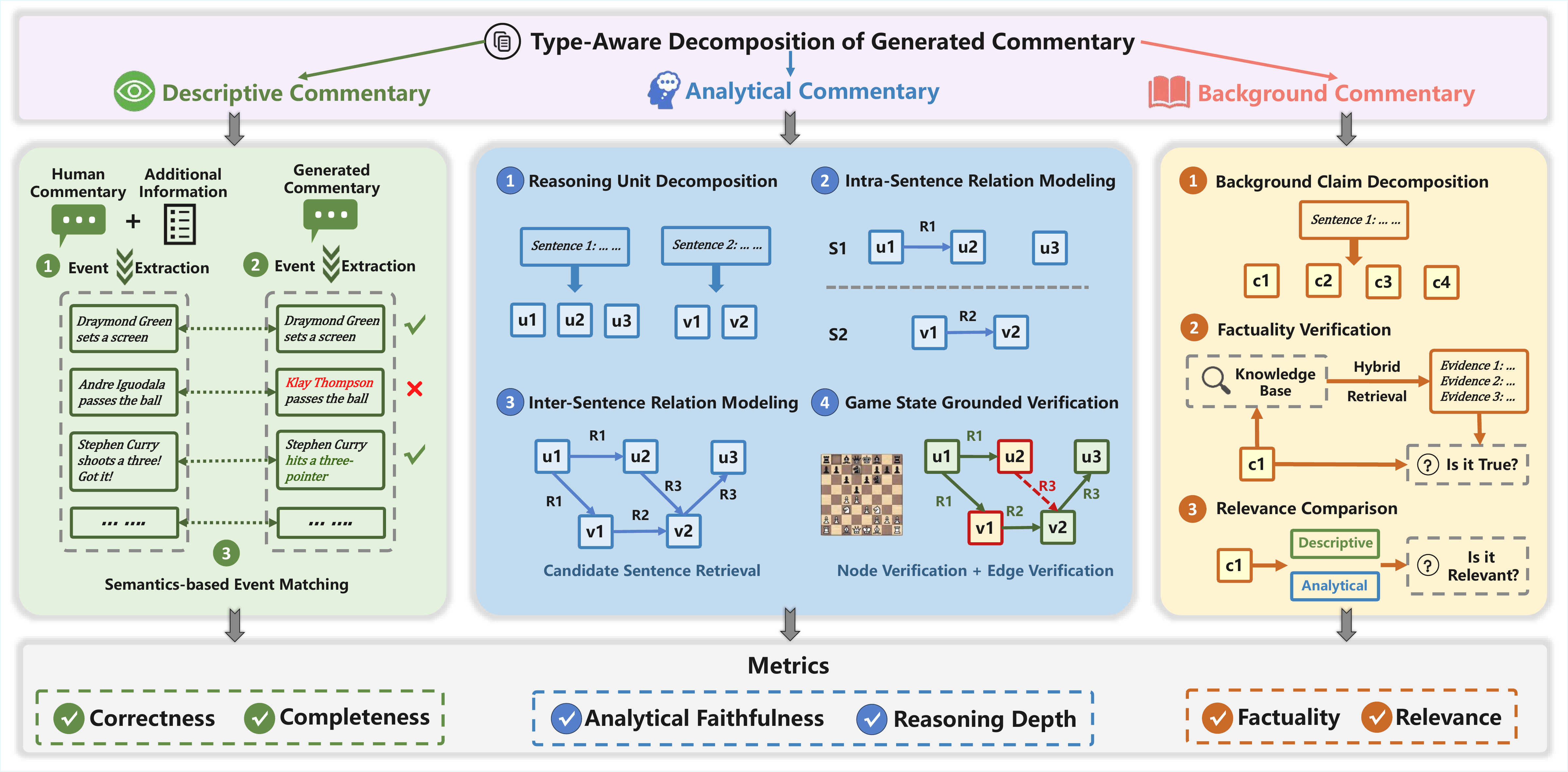}
    \vspace{-4.5ex}
    \caption{\textbf{Overview of TACE.}
TACE evaluates descriptive, analytical, and background commentary through three type-specific pipelines. The resulting metrics assess the description correctness and completeness, reasoning faithfulness and depth, and background factuality and relevance.}
    \label{eval}
    \vspace{-1.3ex}
\end{figure*}

Game commentary is open-ended and contains different functional types, so a single holistic score is insufficient. We propose TACE, which evaluates descriptive, analytical, and background commentary with criteria tailored to each type. Figure~\ref{eval} summarizes the framework: commentary is decomposed by type and evaluated through three branches. We next describe the three pipelines and their metrics.

\subsection{Descriptive Commentary Evaluation}

Descriptive commentary focuses on \emph{what is happening}. Therefore, instead of comparing surface text with human references, we measure whether the generated commentary accurately describes and covers salient events in the current game context.

For each instance, we construct a reference event set $E_{\mathrm{ref}}$ by extracting key actions from the descriptive units of the human commentary. To improve extraction accuracy, we incorporate domain-specific signals: dataset event labels for sports, explicit game-state representations for board games, and significant API changes for LoL. We use GPT-5.4 to extract reference events, GPT-5.4-mini to extract events from generated commentary, and GLM-4.7 to semantically align generated events with the reference event set. This extraction and matching pipeline is validated through human annotation in Section~\ref{sec4.2}.

Given the aligned event sets, we define two complementary metrics. \textbf{Correctness}, computed as $|E_{\mathrm{gen}} \cap E_{\mathrm{ref}}| / |E_{\mathrm{gen}}|$, measures the proportion of generated events supported by the reference set. \textbf{Completeness}, computed as $|E_{\mathrm{gen}} \cap E_{\mathrm{ref}}| / |E_{\mathrm{ref}}|$, measures the fraction of reference events covered by the generated commentary. Correctness penalizes unsupported event descriptions, while completeness penalizes missing salient events.

\subsection{Analytical Commentary Evaluation}

Analytical commentary interprets move quality, state significance, or future trajectories, and its evaluation should prioritize the correctness of the reasoning rather than surface fluency. Open-ended reasoning evaluation is often simplified by turning the output into option selection, or by choosing tasks with a unique answer to limit open-endedness~\citep{wen2025chessqa, yao2025mmreason, zhang2025turnbench}. This setting does not fit AI-GGC: one game state may support multiple valid commentaries, and especially for analytical commentary, there is neither a single correct answer nor a single correct reasoning process. At the same time, evaluating lengthy analytical commentary holistically is also unreliable, since intermediate steps in a reasoning chain may be partially correct, unsupported, or internally inconsistent~\citep{lightman2023let}.

We therefore need to verify whether each piece of articulated reasoning in the commentary is justified under the current game state. To achieve this, we decompose analytical text into granular, explicitly verifiable units. Specifically, each analytical commentary segment is converted into a directed graph, where nodes are \emph{reasoning units}, the smallest logically meaningful propositions in the commentary, and edges denote predefined logical relations expressed in the text. We first decompose each analytical sentence into reasoning units, and then construct relations among them in two stages. \textbf{Intra-sentence relation modeling} considers candidate reasoning-unit pairs within the same sentence and determines whether an explicit logical relation is present. \textbf{Inter-sentence relation modeling} extends this process across sentences within the same analytical segment: we first retrieve a restricted set of candidate sentences using embedding and keyword similarity, and then construct relations across the paired sentences. Importantly, graph construction is purely text-based: no game-state or auxiliary information is provided. The resulting graph therefore represents the reasoning structure verbalized by the commentary itself. We then perform game-state-grounded verification on the constructed graph. At the node level, we assess whether each reasoning unit is supported by the corresponding game state. At the edge level, we assess whether the logical relation between two reasoning units is justified under the same game state. During verification, an edge is considered valid only when both endpoint reasoning units are supported and the expressed relation is justified. This decomposition mitigates the difficulty of LLM-as-a-Judge~\citep{gu2024survey}, enabling fine-grained assessment of strategic reasoning correctness instead of relying on holistic plausibility. Both relation modeling and graph verification are implemented using GPT-5.4-mini, and their accuracy is validated in Section~\ref{sec4.2}.

Accordingly, we define two metrics for analytical commentary. \textbf{Analytical Faithfulness} measures the fraction of reasoning units and relations that pass game-state-grounded verification. \textbf{Reasoning Depth} measures the longest verified reasoning chain in the constructed graph, normalized by the largest value achieved by any evaluated model on the same instance. Together, they capture whether analytical reasoning is grounded and how far valid reasoning develops.

\subsection{Background Commentary Evaluation}

Background commentary provides contextual or historical information beyond immediate game events. We evaluate it along two dimensions: \textbf{Factuality} and \textbf{Relevance}. For granular evaluation, we first decompose each background segment into \emph{background claims}, defined as factual assertions distilled from the commentary.

\vspace{-0.2ex}
\textbf{Factuality} assesses the objective accuracy of each background claim. Rather than relying solely on the evaluator's parametric knowledge, we support this judgment with an offline structured knowledge base constructed through iterative retrieval. This knowledge base provides external evidence for the relevant game, team, player, or match context. For each claim, we retrieve an evidence set using a hybrid strategy that combines indexed retrieval over structured fields, such as matches, teams, and players, with semantic retrieval based on claim--evidence similarity. The retrieved evidence is then provided to the evaluator for evidence-grounded factuality judgment, and the final factuality score is computed by averaging claim-level judgments over the background segment. \textbf{Relevance} evaluates whether a background claim aligns with the immediate context of the commentary. Specifically, we compare each claim against the game state and the remaining portions of the segment, including descriptive and analytical components, to determine whether the information is contextually integrated or unrelated. The relevance score is similarly derived by averaging claim-level judgments. Appendices~\ref{rubric1} and~\ref{rubric2} provide the detailed rubrics and evidence retrieval protocols.

\begin{table*}[t]
\vspace{-0.8cm}
\centering
\caption{\textbf{Validation of TACE.} We report run-to-run reliability and human validation for LLM-mediated components.}
\label{tab:tace_validation}
\vspace{-0.8ex}
\begin{subtable}[t]{0.608\textwidth}
\centering
\caption{Reliability}
\vspace{-0.5ex}
\label{tab:tace_reliability}
\scriptsize
\resizebox{\linewidth}{!}{
\setlength{\tabcolsep}{2pt}
\begin{tabular}{llccc}
\toprule
\textbf{Type} & \textbf{Metric} & \textbf{Mean $\pm$ SD} & \textbf{CV(\%)} & \textbf{Spearman $\rho$} \\
\midrule
\multirow{2}{*}{Desc.}
& Correctness & 0.2312$\pm$0.0025 & 1.08 & 0.918 \\
& Completeness & 0.3215$\pm$0.0026 & 0.81 & 0.965 \\
\midrule
\multirow{2}{*}{Analyt.}
& Faithfulness & 0.3605$\pm$0.0112 & 3.11 & 0.905 \\
& Reasoning Depth & 0.0465$\pm$0.0016 & 3.44 & 0.856 \\
\midrule
\multirow{2}{*}{Back.}
& Factuality & 0.7358$\pm$0.0132 & 1.79 & 0.918 \\
& Relevance & 0.9185$\pm$0.0202 & 2.20 & 0.868 \\
\bottomrule
\end{tabular}
}
\end{subtable}
\hfill
\begin{subtable}[t]{0.372\textwidth}
\centering
\caption{Component accuracy}
\vspace{-0.5ex}
\label{tab:tace_human_validation}
\scriptsize
\resizebox{\linewidth}{!}{
\setlength{\tabcolsep}{2pt}
\begin{tabular}{llc}
\toprule
\textbf{Type} & \textbf{Validated Step} & \textbf{Acc.(\%)} \\
\midrule
\multirow{2}{*}{Desc.}
& Event Extraction & 91.16 \\
& Event Matching & 94.73 \\
\midrule
\multirow{2}{*}{Analyt.}
& Relation Modeling & 93.40 \\
& Relation Verification & 81.43 \\
\midrule
\multirow{2}{*}{Back.}
& Factuality Judgment & 92.86 \\
& Relevance Judgment & 94.44 \\
\bottomrule
\end{tabular}
}
\end{subtable}
\vspace{-1.5ex}
\end{table*}

\section{Validation of TACE}

We validate TACE along two axes: reliability and human agreement. Reliability measures run-to-run score variation and ranking consistency across repeated evaluations, while human validation checks whether the intermediate LLM-mediated decisions align with human annotations.

\subsection{Reliability of TACE}
\label{sec4.1}

To assess the reliability of TACE, we repeat the full evaluation pipeline three times on a held-out subset covering 6 models, 5 games, and 20 randomly sampled cases for each model-game pair. For each of the six TACE metrics, we report the mean and standard deviation across repeated runs, together with the coefficient of variation (CV) to quantify relative score fluctuation. Since benchmark conclusions also depend on model ordering, we further compute pairwise Spearman rank correlations across runs to assess ranking stability.
\vspace{-0.1ex}

As shown in Table~\ref{tab:tace_reliability}, TACE exhibits low run-to-run variation across all six metrics. The CV values remain below 3.5\%, indicating strong score-level reproducibility. Descriptive metrics show particularly small variation, while analytical metrics exhibit slightly higher but still limited fluctuation, as expected given the additional complexity of reasoning-graph construction and verification. Ranking consistency is also high, with Spearman correlations ranging from 0.856 to 0.965 across metrics. These results suggest that TACE provides stable type-specific evaluation signals for model comparison. We additionally report Spearman rank correlations in Appendix~\ref{cc}, where GPT-5.4-mini is replaced by Gemini-3-Flash as the judge model.

\subsection{Human Validation}
\label{sec4.2}

To assess the validity of TACE, we conduct human validation on the LLM-mediated components used in its evaluation pipeline. We first validate the accuracy of LLM-based type-aware decomposition. For each game, we randomly sample 100 decomposed commentary and manually label their types. The results, reported in Table~\ref{tab:type_decomposition_acc}, show consistently high decomposition accuracy across all games, with overall accuracy exceeding 95\%.

\vspace{-0.2ex}
Then, we validate whether each LLM-mediated step in TACE pipeline are consistent with human annotations. We use a stratified random sample covering all 6 evaluated models and 5 games, with three instances sampled for each model-game pair. This results in 90 annotated instances and more than 3,700 intermediate validation items, including event extraction outputs, event-matching decisions, reasoning-graph relations, verification judgments, and background-claim judgments. For descriptive commentary, we validate event extraction and semantic event matching. For analytical commentary, we validate whether the constructed relations accurately reflect the logical structure expressed in the text, and whether relation verification correctly judges the validity of reasoning connections with respect to the game state. For background commentary, we assess the accuracy of factuality and relevance judgments for extracted background claims. As shown in Table~\ref{tab:tace_human_validation}, the LLM-mediated components achieve high agreement with human annotations across all components, with most validated steps exceeding 90\% accuracy. These results support the validity of the intermediate automatic decisions used by TACE.



\section{Experiments}
\vspace{-0.2ex}

\begin{figure*}[t]
\centering
\vspace{-0.8cm}
\small
\setlength{\tabcolsep}{7pt}
\begin{tabular}{lcccccc}
\toprule
\multirow{2}{*}{\textbf{Model}} 
& \multicolumn{2}{c}{\textbf{Descriptive}} 
& \multicolumn{2}{c}{\textbf{Analytical}} 
& \multicolumn{2}{c}{\textbf{Background}} \\
\cmidrule(lr){2-3} \cmidrule(lr){4-5} \cmidrule(lr){6-7}
& \textbf{Correctness} 
& \textbf{Completeness} 
& \textbf{Faithfulness}
& \textbf{Depth}
& \textbf{Factuality} 
& \textbf{Relevance} \\
\midrule
Gemini-3.1-Pro-Preview & 0.206 & 0.499 & 0.258 & 0.171 & 0.691 & 0.865 \\
GPT-5.4                & \textbf{0.281} & \textbf{0.610} & \textbf{0.439} & \textbf{0.318} & 0.655 & 0.922 \\
Claude-Sonnet-4.5      & 0.243 & 0.533 & 0.300 & 0.196 & 0.661 & 0.911 \\
Qwen3-VL-32B           & 0.217 & 0.467 & 0.163 & 0.197 & 0.671 & \textbf{0.927} \\
GLM-4.6V               & 0.220 & 0.282 & 0.190 & 0.185 & 0.682 & \textbf{0.927} \\
InternVL3.5-38B        & 0.196 & 0.445 & 0.124 & 0.114 & \textbf{0.734} & 0.866 \\
\midrule
\textbf{Average}       & 0.227 & 0.473 & 0.246 & 0.196 & 0.683 & 0.903 \\
\bottomrule
\end{tabular}

\vspace{1.0ex}

\makebox[\linewidth][r]{%
  \includegraphics[width=1\linewidth]{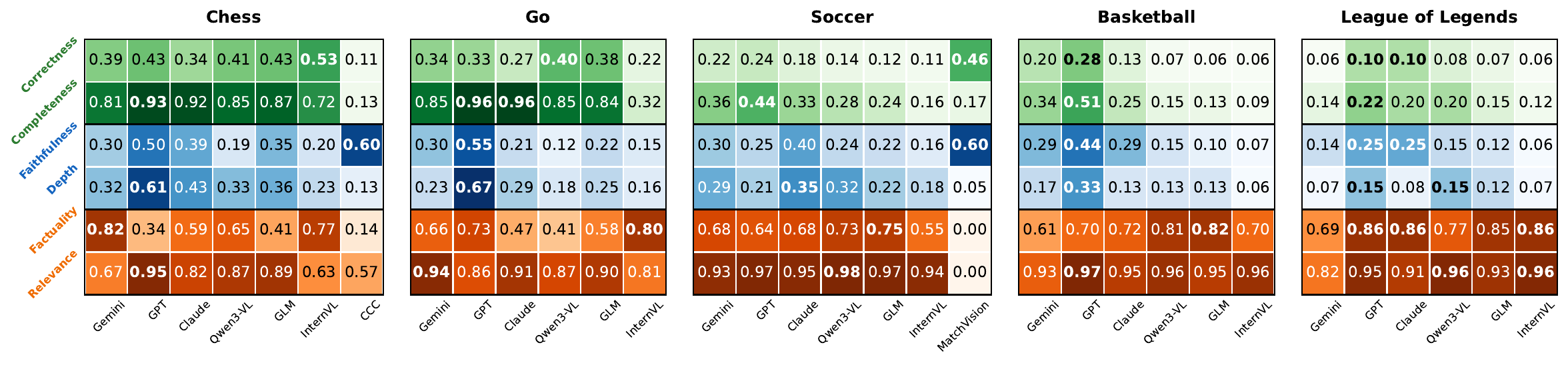}%
}
\vspace{-4ex}
\caption{\textbf{Main benchmark results on \textsc{GameCommBench}.}
The top table reports aggregated model scores for the six TACE metrics, with the average row summarizing metric-wise performance across evaluated models. The bottom heatmaps show game-wise capability profiles across models and metrics, revealing cross-game and type-specific performance patterns.}
\label{fig:main_results}
\vspace{-1ex}
\end{figure*}

Based on \textsc{GameCommBench} and TACE, we evaluate representative VLM commentators across games and commentary types, then analyze capability imbalances and the effects of model scale and reasoning mode.

\subsection{Evaluation Setup}

We evaluate six representative state-of-the-art VLMs: three closed-source models, Gemini-3.1-Pro-Preview, GPT-5.4, and Claude-Sonnet-4.5, and three open-weight models, Qwen3-VL-32B, GLM-4.6V, and InternVL3.5-38B. All models are evaluated on the full \textsc{GameCommBench} benchmark. Each instance includes basic player/team and game information. For board games, models receive symbolic game states with local move history, including the current move and the preceding four moves, corresponding to two full turns. For sports and esports, models process video inputs sampled at 4 FPS while preserving the original resolution; the prompt also specifies clip duration and asks the model to generate commentary appropriate for that duration. For esports, we additionally provide timestamp-aligned game-state metadata derived from the game API. Human commentary is not provided during generation.

\vspace{-0.2ex}
All models use a shared commentary-generation template, with minor modality-specific adaptations for symbolic states, videos, and API-based metadata. We set temperature to 0 and the maximum output length to 1024 tokens, with one response generated per instance. To ensure comparable inference conditions, explicit reasoning modes are disabled when supported; otherwise, we use the minimal reasoning setting. Other decoding parameters are left at their default values.

\vspace{-0.2ex}
We also evaluate two existing game-specific AI-GGC systems with TACE: Concept-guided Chess Commentary generation (CCC)~\citep{kim2025bridging} for chess and MatchVision~\citep{rao2025towards} for soccer. Their results are included in the heatmaps in Figure~\ref{fig:main_results}, enabling comparison between general-purpose VLM commentators and task-specific AI-GGC systems.

\subsection{Main Benchmark Results}

Figure~\ref{fig:main_results} summarizes the overall and game-wise performance of the six VLM commentators. The results reveal substantial capability imbalance across both games and commentary types. No model consistently dominates all metrics: GPT-5.4 performs best on descriptive and analytical metrics, whereas other models remain competitive on background factuality or relevance. This confirms that AI-GGC capability is not one-dimensional and should not be collapsed into a single holistic score.

\begin{table*}[t]
\vspace{-0.8cm}
\centering
\caption{\textbf{Effects of model scale and reasoning mode.}
Left: comparison of open-weight models with different sizes. Right: comparison between minimal-reasoning and thinking settings on board games.}
\vspace{-0.1cm}
\label{tab:scale_thinking}

\begin{subtable}[t]{0.68\textwidth}
\centering
\caption{Effect of Model Scale}
\label{tab:scale}
\scriptsize
\setlength{\tabcolsep}{2pt}
\resizebox{\linewidth}{!}{
\begin{tabular}{llcccccc}
\toprule
\multirow{2}{*}{\textbf{Model}} 
& \multirow{2}{*}{\textbf{Size}} 
& \multicolumn{2}{c}{\textbf{Descriptive}} 
& \multicolumn{2}{c}{\textbf{Analytical}} 
& \multicolumn{2}{c}{\textbf{Background}} \\
\cmidrule(lr){3-4} \cmidrule(lr){5-6} \cmidrule(lr){7-8}
& & \textbf{Correctness} 
& \textbf{Completeness} 
& \textbf{Faithfulness}
& \textbf{Depth}
& \textbf{Factuality}
& \textbf{Relevance} \\
\midrule
\multirow{2}{*}{Qwen3-VL} & 32B & 0.252 & 0.533 & 0.169 & 0.232 & 0.647 & 0.920 \\
 & 8B & 0.260 & 0.500 & 0.111 & 0.184 & 0.566 & 0.916 \\
\midrule
\multirow{2}{*}{InternVL3.5} & 38B  & 0.229 & 0.322 & 0.148 & 0.151 & 0.704 & 0.835 \\
 & 8B   & 0.216 & 0.319 & 0.137 & 0.188 & 0.580 & 0.887 \\
\midrule
\multirow{2}{*}{GLM-4.6V} & 106B & 0.247 & 0.520 & 0.227 & 0.231 & 0.639 & 0.926 \\
 & 9B & 0.257 & 0.492 & 0.160 & 0.112 & 0.511 & 0.923 \\
\bottomrule
\end{tabular}
}
\end{subtable}
\hfill
\begin{subtable}[t]{0.30\textwidth}
\centering
\caption{Effect of Reasoning Mode}
\label{tab:think}
\scriptsize
\setlength{\tabcolsep}{2pt}
\resizebox{\linewidth}{!}{
\begin{tabular}{lccc}
\toprule
\multirow{2}{*}{\textbf{Model}} 
& \multirow{2}{*}{\makecell{\textbf{Thinking}\\\textbf{Level}}}
& \multicolumn{2}{c}{\textbf{Analytical}} \\
\cmidrule(lr){3-4}
& & \textbf{Faithfulness}
& \textbf{Depth} \\
\midrule
\multirow{2}{*}{Gemini} & Min  & 0.206 & 0.271 \\
 & Max & 0.532 & 0.451 \\
\midrule
\multirow{2}{*}{GPT}    & Min  & 0.250 & 0.427 \\
 & Max & 0.364 & 0.445 \\
\midrule
\multirow{2}{*}{Claude} & Min  & 0.235 & 0.394 \\
 & Max & 0.246 & 0.357 \\
\bottomrule
\end{tabular}
}
\end{subtable}

\end{table*}

For descriptive commentary, models show a clear gap between completeness and correctness. The average completeness reaches 0.473, while correctness is only 0.227, suggesting that VLMs often generate more event descriptions than can be directly supported by the reference signals. This pattern indicates a verbosity--grounding mismatch: models often describe plausible events, but many of these descriptions are not supported by the reference events or available game-state signals. In board games, this issue is partly mitigated by the compact symbolic input, and lower correctness may also reflect the fact that human commentators do not mention every objectively valid move-level event. However, the same pattern becomes much more problematic in video-based games. For sports and esports, both correctness and completeness drop sharply, as shown in Figure~\ref{fig:main_results}, indicating that current VLMs not only miss salient events but also generate unsupported descriptions in fast-paced multi-agent visual scenes. This exposes live observation as a major bottleneck for AI-GGC in dynamic video settings.

Analytical commentary remains a major bottleneck. The average analytical faithfulness is 0.246 and the average reasoning depth is 0.196, showing that most models have difficulty producing game-grounded strategic reasoning. GPT-5.4 achieves the best analytical performance, but the absolute scores remain limited. This indicates that even strong VLMs often produce analyses that are shallow, partially unsupported, or weakly connected to underlying game state. For reference, human analytical commentary shows high faithfulness (0.78) but relatively low depth (0.28), suggesting that expert commentary can be strategically correct while remaining concise, rather than explicitly spelling out long reasoning chains. This distinction also shows why faithfulness and depth should be interpreted as complementary analytical signals.

Background commentary shows a different capability profile. Models achieve high relevance, reaching 0.903 on average, while factuality remains moderate at 0.683. Notably, open-weight models achieve the strongest scores on both background metrics, unlike the pattern observed for descriptive and analytical metrics. The two AI-GGC systems show limited performance on background metrics, indicating that historical recall remains underexplored in existing AI-GGC work. Overall, current VLMs can usually generate background information that is topically connected to the commentary context, likely benefiting from parametric knowledge and the basic game information provided in the prompt. However, there is still room for improvement for the factuality, suggesting that background commentary may benefit from retrieval-augmented or evidence-grounded generation mechanisms that provide verified player, team, match, or historical information.

\subsection{Scale and Reasoning Mode}

We further analyze two factors that may affect commentary capability: model scale and reasoning mode. For model scale, we compare the three open-weight models in the main result with their lighter variants on the full board-game and sports subsets. For reasoning mode, we evaluate the three closed-source models on a randomly sampled quarter of the board-game turns, comparing the minimal-reasoning setting used in the main experiments with the maximum thinking setting supported by each model.

As shown in Table~\ref{tab:scale}, increasing model scale does not lead to uniform improvements in descriptive commentary. Larger models generally achieve slightly higher completeness, but their correctness remains similar or even decreases. This suggests that scaling makes models describe more events, but not a higher proportion of supported events. This pattern is consistent with the main benchmark results, where descriptive correctness remains a key bottleneck, especially for video-based games.

The effect of scale is more evident for analytical and background capabilities. Larger models generally improve analytical faithfulness, indicating better grounding of strategic statements in the game state. However, reasoning depth does not increase consistently across model families, suggesting that scale alone does not guarantee longer valid reasoning chains. For background commentary, factuality shows the clearest scaling trend, with larger variants consistently achieving higher factuality scores. This may reflect stronger parametric knowledge, better entity-level understanding, or more effective use of the basic game information provided in the prompt. In contrast, relevance is already high for most models and is less sensitive to scale.

As shown in Table~\ref{tab:think}, reasoning mode has a clear positive effect on analytical commentary. Across the three closed-source models, increasing the thinking level improves five out of six analytical scores. The gains are particularly strong for Gemini-3.1-Pro-Preview, whose faithfulness and reasoning depth both increase substantially, making it the strongest model in this board-game analytical setting. GPT-5.4 also benefits from the higher thinking setting, with a clear improvement in faithfulness and a smaller gain in reasoning depth. Claude-Sonnet-4.5 shows a weaker response: faithfulness slightly improves, while reasoning depth decreases modestly. Overall, these results indicate that stronger thinking settings can improve analytical commentary, especially analytical faithfulness, although their effect on reasoning depth remains more model-dependent.

\vspace{-0.2ex}
\section{Conclusion}
\vspace{-0.2ex}



We introduced \textsc{GameCommBench}, a unified benchmark for AI-GGC across board games, sports, and esports, together with TACE, a type-aware commentary evaluation framework for descriptive, analytical, and background commentary. Our validation shows that TACE provides stable and human-aligned evaluation signals. Experiments with representative AI commentators reveal non-uniform capability profiles across games and commentary types, with grounded game state description and faithful strategic reasoning remaining key bottlenecks. Further analyses show that model scale and reasoning mode bring uneven gains across commentary capabilities. Together, \textsc{GameCommBench} and TACE support more comparable and interpretable evaluation of AI-GGC.

\section*{Limitations}

\textsc{GameCommBench} is designed to cover representative AI-GGC settings across board games, sports, and esports, but it does not exhaust all game genres, languages, or broadcasting styles. The current benchmark focuses on English text commentary generation and evaluation over five games with different state representations; it does not address speech-based commentary, voice delivery, or other multimodal presentation forms of game broadcasting. Extensions to additional languages, regional commentary styles, game communities, and speech or multimodal commentary settings may require new data collection and domain-specific processing. Finally, full multimodal evaluation across multiple games and VLMs is costly, especially with closed-source APIs.

\section*{Ethical Considerations}
\label{ethics}

This work studies benchmark construction and evaluation for AI-GGC. Generated commentary may contain errors, such as inaccurate descriptions or unsupported interpretations, which could mislead viewers if used directly in public-facing commentary systems, especially in text-only live updates where viewers rely primarily on the generated text. In addition, \textsc{GameCommBench} is constructed from game records, videos, transcripts, API-derived metadata, and related contextual information. These sources may contain public names of players, teams, matches, or commentators that are necessary for game commentary and background evaluation; however, we do not collect private or sensitive personal information or intentionally include offensive content. The benchmark is constructed from public or previously released game materials, and no separate consent was obtained from players, teams, or commentators. Users of the benchmark should respect the licenses, terms of use, and attribution requirements of the original data sources.


\bibliography{custom}

@article{zheng2025multimodal,
  title={From Multimodal Perception to Strategic Reasoning: A Survey on AI-Generated Game Commentary},
  author={Zheng, Qirui and Wang, Xingbo and Cheng, Keyuan and Ali, Muhammad Asif and Lu, Yunlong and Li, Wenxin},
  journal={arXiv preprint arXiv:2506.17294},
  year={2025}
}

@article{behrens2022you,
  title={You'll never want to watch alone: The effect of displaying in-stadium social atmospherics on media consumers' responses to new sport leagues across different types of media},
  author={Behrens, Anton and Uhrich, Sebastian},
  journal={European Sport Management Quarterly},
  volume={22},
  number={1},
  pages={120--138},
  year={2022},
  publisher={Taylor \& Francis}
}

@inproceedings{giancola2018soccernet,
  title={Soccernet: A scalable dataset for action spotting in soccer videos},
  author={Giancola, Silvio and Amine, Mohieddine and Dghaily, Tarek and Ghanem, Bernard},
  booktitle={Proceedings of the IEEE conference on computer vision and pattern recognition workshops},
  pages={1711--1721},
  year={2018}
}

@inproceedings{wu2022sports,
  title={Sports video analysis on large-scale data},
  author={Wu, Dekun and Zhao, He and Bao, Xingce and Wildes, Richard P},
  booktitle={European conference on computer vision},
  pages={19--36},
  year={2022},
  organization={Springer}
}

@inproceedings{jhamtani2018learning,
  title={Learning to generate move-by-move commentary for chess games from large-scale social forum data},
  author={Jhamtani, Harsh and Gangal, Varun and Hovy, Eduard and Neubig, Graham and Berg-Kirkpatrick, Taylor},
  booktitle={Proceedings of the 56th Annual Meeting of the Association for Computational Linguistics (Volume 1: Long Papers)},
  pages={1661--1671},
  year={2018}
}

@article{zhang2024game,
  title={Game-MUG: Multimodal oriented game situation understanding and commentary generation dataset},
  author={Zhang, Zhihao and Cao, Feiqi and Mo, Yingbin and Zhang, Yiran and Poon, Josiah and Han, Caren},
  journal={arXiv preprint arXiv:2404.19175},
  year={2024}
}

@inproceedings{kim2025bridging,
  title={Bridging the gap between expert and language models: Concept-guided chess commentary generation and evaluation},
  author={Kim, Jaechang and Goh, Jinmin and Hwang, Inseok and Cho, Jaewoong and Ok, Jungseul},
  booktitle={Proceedings of the 2025 Conference of the Nations of the Americas Chapter of the Association for Computational Linguistics: Human Language Technologies (Volume 1: Long Papers)},
  pages={9497--9516},
  year={2025}
}

@article{gu2024survey,
  title={A survey on llm-as-a-judge},
  author={Gu, Jiawei and Jiang, Xuhui and Shi, Zhichao and Tan, Hexiang and Zhai, Xuehao and Xu, Chengjin and Li, Wei and Shen, Yinghan and Ma, Shengjie and Liu, Honghao and others},
  journal={The Innovation},
  year={2024},
  publisher={Elsevier}
}

@inproceedings{lightman2023let,
  title={Let's verify step by step},
  author={Lightman, Hunter and Kosaraju, Vineet and Burda, Yuri and Edwards, Harrison and Baker, Bowen and Lee, Teddy and Leike, Jan and Schulman, John and Sutskever, Ilya and Cobbe, Karl},
  booktitle={The twelfth international conference on learning representations},
  year={2023}
}

@inproceedings{rao2025towards,
  title={Towards universal soccer video understanding},
  author={Rao, Jiayuan and Wu, Haoning and Jiang, Hao and Zhang, Ya and Wang, Yanfeng and Xie, Weidi},
  booktitle={Proceedings of the Computer Vision and Pattern Recognition Conference},
  pages={8384--8394},
  year={2025}
}

@article{bain2023whisperx,
  title={Whisperx: Time-accurate speech transcription of long-form audio},
  author={Bain, Max and Huh, Jaesung and Han, Tengda and Zisserman, Andrew},
  journal={arXiv preprint arXiv:2303.00747},
  year={2023}
}

@article{lee2022improving,
  title={Improving chess commentaries by combining language models with symbolic reasoning engines},
  author={Lee, Andrew and Wu, David and Dinan, Emily and Lewis, Mike},
  journal={arXiv preprint arXiv:2212.08195},
  year={2022}
}

@inproceedings{you2025timesoccer,
  title={Timesoccer: An end-to-end multimodal large language model for soccer commentary generation},
  author={You, Ling and Huang, Wenxuan and Xie, Xinni and Wei, Xiangyi and Li, Bangyan and Lin, Shaohui and Li, Yang and Wang, Changbo},
  booktitle={Proceedings of the 33rd ACM International Conference on Multimedia},
  pages={3418--3427},
  year={2025}
}

@inproceedings{tomlin2022understanding,
  title={Understanding game-playing agents with natural language annotations},
  author={Tomlin, Nicholas and He, Andre and Klein, Dan},
  booktitle={Proceedings of the 60th Annual Meeting of the Association for Computational Linguistics (Volume 2: Short Papers)},
  pages={797--807},
  year={2022}
}

@inproceedings{liu2023g,
  title={G-eval: NLG evaluation using gpt-4 with better human alignment},
  author={Liu, Yang and Iter, Dan and Xu, Yichong and Wang, Shuohang and Xu, Ruochen and Zhu, Chenguang},
  booktitle={Proceedings of the 2023 conference on empirical methods in natural language processing},
  pages={2511--2522},
  year={2023}
}

@inproceedings{deliege2021soccernet,
  title={Soccernet-v2: A dataset and benchmarks for holistic understanding of broadcast soccer videos},
  author={Deliege, Adrien and Cioppa, Anthony and Giancola, Silvio and Seikavandi, Meisam J and Dueholm, Jacob V and Nasrollahi, Kamal and Ghanem, Bernard and Moeslund, Thomas B and Van Droogenbroeck, Marc},
  booktitle={Proceedings of the IEEE/CVF conference on computer vision and pattern recognition},
  pages={4508--4519},
  year={2021}
}

@inproceedings{yu2018fine,
  title={Fine-grained video captioning for sports narrative},
  author={Yu, Huanyu and Cheng, Shuo and Ni, Bingbing and Wang, Minsi and Zhang, Jian and Yang, Xiaokang},
  booktitle={Proceedings of the IEEE Conference on Computer Vision and Pattern Recognition},
  pages={6006--6015},
  year={2018}
}

@article{yan2019fine,
  title={Fine-grained video captioning via graph-based multi-granularity interaction learning},
  author={Yan, Yichao and Zhuang, Ning and Ni, Bingbing and Zhang, Jian and Xu, Minghao and Zhang, Qiang and Zhang, Zheng and Cheng, Shuo and Tian, Qi and Xu, Yi and others},
  journal={IEEE transactions on pattern analysis and machine intelligence},
  volume={44},
  number={2},
  pages={666--683},
  year={2019},
  publisher={IEEE}
}

@inproceedings{wang2025commentary,
  title={Commentary Generation from Multimodal Game Data for Esports Moments in Multiplayer Strategy Games},
  author={Wang, Zihan and Yoshinaga, Naoki},
  booktitle={Proceedings of the 14th International Joint Conference on Natural Language Processing and the 4th Conference of the Asia-Pacific Chapter of the Association for Computational Linguistics},
  pages={1795--1807},
  year={2025}
}

@article{wen2025chessqa,
  title={ChessQA: Evaluating Large Language Models for Chess Understanding},
  author={Wen, Qianfeng and Tang, Zhenwei and Anderson, Ashton},
  journal={arXiv preprint arXiv:2510.23948},
  year={2025}
}

@article{zhang2025turnbench,
  title={TurnBench-MS: A benchmark for evaluating multi-turn, multi-step reasoning in large language models},
  author={Zhang, Yiran and Wang, Mo and Li, Xiaoyang and Ren, Kaixuan and Zhu, Chencheng and Naseem, Usman},
  journal={Findings of the Association for Computational Linguistics: EMNLP},
  pages={19892--19924},
  year={2025}
}

@inproceedings{yao2025mmreason,
  title={Mmreason: An open-ended multi-modal multi-step reasoning benchmark for mllms toward agi},
  author={Yao, Huanjin and Huang, Jiaxing and Qiu, Yawen and Chen, Michael K and Liu, Wenzheng and Zhang, Wei and Zeng, Wenjie and Zhang, Xikun and Zhang, Jingyi and Song, Yuxin and others},
  booktitle={Proceedings of the IEEE/CVF International Conference on Computer Vision},
  pages={273--283},
  year={2025}
}

\appendix

\begin{figure*}[t]
    \centering
    \includegraphics[width=\linewidth]{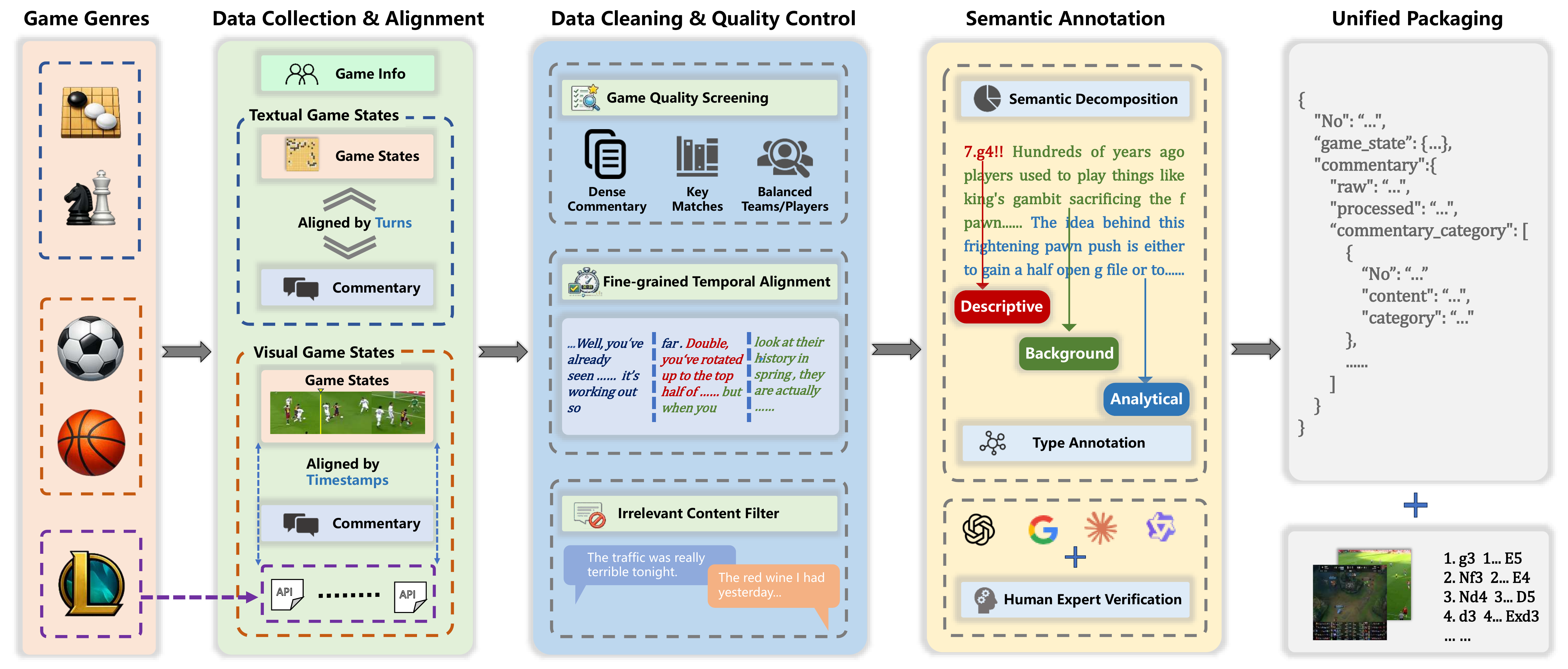}
    \caption{\textbf{Detailed data processing pipeline.} Raw game data are collected and aligned, cleaned through quality control and temporal repair, decomposed into commentary types, and packaged into a unified format.}
\label{fig:pipeline_appendix}
\end{figure*}

\section{Detail of Data Collection and Processing Pipeline}

This appendix provides implementation-level details for the data collection and processing pipeline among the five game genres in \textsc{GameCommBench}. As shown in Figure~\ref{fig:pipeline_appendix}, the main text already describes the high-level collection, cleaning, and packaging strategy; here we focus on the concrete processing steps used to convert raw game state data and commentary into cleaned benchmark instances.

\subsection{Unified record schema and normalization}

The final conversion stage packages heterogeneous samples into a common JSONL interface. Sports records use the complete schema below, and board-game/esports records follow the same semantic contract while using PGN, SGF, or API-derived fields as the game-state carrier.

\begin{PromptCode}
{
  "No": "...",
  "game_state": {
    "modality": "video | pgn | sgf | api+video",
    "storage_file": "...",
    "step": "...",
    "label": "..."
  },
  "background": "...",
  "commentary": {
    "raw_commetary": "...",
    "processed_commentary": "...",
    "commentary_category": [
      {
        "No": "...",
        "content": "...",
        "category": "descriptive | analytical | background"
      }
    ]
  }
}
\end{PromptCode}

Here, \code{raw_commetary} keeps the field name used by the implementation, and category IDs are normalized into \code{descriptive}, \code{analytical}, and \code{background}. For football and basketball, the original video or event ID is replaced by an MD5-based stable identifier, with the original-to-hash mapping retained separately.

\subsection{Game-specific processing workflows}
\label{app:data-processing-workflows}

This subsection merges the collection, alignment, cleaning, and final annotation steps for each game type. The degree of processing differs by source modality: chess is the lightest because the source comments are already attached to PGN plies; Go requires extraction from FoxWQ records plus removal of unusable commentary; football, basketball, and LoL require explicit temporal alignment and cleaning. The LLM prompts used by the released processing code are reproduced after the workflow descriptions.

\subsubsection{Football}

Football starts from raw rows containing a clip caption and ASR-style commentary. The retained rows are tied to externally supplied video clips through the football video manifest, and match-level background is reconstructed from the raw match metadata, including teams, date, venue, referee, and player lists. The processing chain then performs: (i) sentence segmentation and type classification over the transcript with Prompt A.1; (ii) caption-grounded filtering with Prompt A.3, where the caption is treated as the authoritative action description and sentences with ambiguous references, weak relevance, or contradictions are removed; (iii) caption-commentary fusion with Prompt A.4, which creates a unified broadcast-style description from the caption and retained commentary; (iv) a final pass of Prompt A.1 over the fused text; and (v) conversion into the unified schema, including label normalization and MD5-based stable identifiers.

\subsubsection{Basketball}

Basketball follows the same alignment-and-cleaning pattern as football, but its initial alignment is stricter because the raw NSVA-derived caption and the transcript may refer to adjacent plays. The pipeline first applies the relevance gate in Prompt A.5 and only writes caption-commentary pairs whose boxed judgment is \verb|\boxed{true}|. The remaining rows are segmented and classified with the basketball version of the classification prompt in Prompt A.2, filtered with the same retention logic as Prompt A.3 using the basketball-specific phrase ``on-court situations,'' and then fused with the basketball caption-commentary prompt in Prompt A.6. The basketball pipeline additionally runs the quality selector in Prompt A.7 over batches of three merged candidates and keeps the two highest-ranked examples. Final conversion attaches the raw NSVA caption and transcript, the video storage path, the coarse action label, the game background recovered from NBA metadata files, normalized commentary categories, and an MD5-based stable identifier.

\begin{table*}[t]
\centering
\small
\begin{tabular}{L{0.15\linewidth}L{0.24\linewidth}L{0.53\linewidth}}
\toprule
\textbf{Game} & \textbf{Primary alignment} & \textbf{Processing summary} \\
\midrule
Football & Caption/event clip to transcript & Segment and classify transcript, filter against caption, fuse caption and retained commentary, reclassify, then normalize IDs and labels. \\
Basketball & NSVA caption/action clip to transcript & Gate caption-commentary relevance, segment/classify, filter, fuse, select high-quality merged outputs, then normalize with NBA metadata and action labels. \\
Chess & PGN mainline ply & Parse GameKnot HTML into PGN comments; preserve move-attached commentary with only lightweight text normalization. \\
Go & SGF mainline ply & Crawl FoxWQ JSON, extract SGF, discard empty or irrelevant comments, and normalize turn-aligned records. \\
League of Legends & Subtitle window plus API window & Align subtitle segments to clips, aggregate Riot/LoL Esports API windows, and retain significant team/player state changes. \\
\bottomrule
\end{tabular}
\caption{Integrated data-processing workflows for the five game types.}
\label{tab:data-processing-paths}
\end{table*}

\subsubsection{Chess}

Chess uses the lightest data-processing path. Saved GameKnot HTML pages are grouped by game ID, sorted by page index, and parsed into player/event metadata plus move-comment tuples. The converter writes one PGN file per game, preserves comments in PGN braces, removes HTML/script artifacts from comments, normalizes whitespace, escapes PGN tag strings, and stores metadata such as event, site, players, ratings, opening, annotator, game ID, and move count. Since commentary is already attached to moves, no timestamp matching or caption-grounded LLM cleaning is needed. The final records use the PGN file as \code{storage_file}, the ply as \code{step}, and the comment text as the reference commentary after lightweight normalization and label mapping.

\subsubsection{Go}

Go starts from FoxWQ metadata seeds. The crawler downloads each referenced JSON payload, and the converter extracts the \code{chess} field into an SGF file named by \code{chessid}. Empty or malformed JSON entries are skipped. Compared with chess, Go needs an additional cleanup step because extracted records can contain empty, duplicated, or off-topic comments; these are removed before the retained move-comment pairs are normalized into the released schema. SGF coordinates are kept as the source representation, while later state construction can convert them into standard Go coordinates and render the board before the target move.

\subsubsection{League of Legends}

LoL combines subtitle/VOD alignment with API-derived state extraction. Subtitle files are parsed into timestamped segments after HTML tag removal, whitespace normalization, and optional consecutive-duplicate removal. For Match-V5 data, events are aggregated in fixed time windows, the most-overlapping subtitle sentence is selected as reference commentary, and each row stores the event window, linearized event summaries, team K/D/A and gold, patch, game date, and a stable MD5 row ID. For LoL Esports live-stat data, the acquisition script resolves match and game IDs from team names, game numbers, league schedules, and event details, fetches live-stat windows, and extracts team/player snapshots around the target timestamp. When start/end snapshots are available, significant API changes are retained as structured context through \code{step_boundary_change}; this records boundary-level changes such as team gold, team kills, and player-level metric changes. 

\section{Candidate Commentary Generation}
\label{app:candidate-generation}

Candidate commentaries are generated without access to the human reference commentary. Each generator reads one JSONL record, constructs the game-specific input context, calls the target model endpoint, and writes a JSONL output with \texttt{No} and \texttt{commentary\_by\_\{model\}}.

For football and basketball, videos are sampled at 4 FPS and the sampled frames are interleaved with timestamp text while preserving the original frame resolution used in the run. For chess and Go, the scripts parse the source game files and pass symbolic state rather than visual frames: chess reconstructs FEN before/after the target move together with SAN/UCI notation and recent move history, while Go renders the board before the target move and supplies the current coordinate plus recent moves. League of Legends uses video input together with match background and, when available, the \texttt{step\_boundary\_change} API summary describing team/player state transitions. Across all reported generation runs, temperature is set to 0 and reasoning or thinking modes are disabled when supported.

\begin{table*}[t]
\centering
\small
\begin{tabular}{@{}L{0.12\linewidth}L{0.30\linewidth}L{0.20\linewidth}L{0.25\linewidth}@{}}
\toprule
\textbf{Game} & \textbf{Input supplied to model} & \textbf{Decoding policy} & \textbf{State construction} \\
\midrule
Football & Timestamped video frames, match background, clip duration & temperature 0; thinking disabled & 4 FPS visual sampling with original frame resolution. \\
Basketball & Timestamped video frames, match background, clip duration & temperature 0; thinking disabled & 4 FPS visual sampling with original frame resolution. \\
Chess & FEN before/after, SAN, UCI, move string, recent moves, background & temperature 0; thinking disabled & PGN mainline parsing with recent move history. \\
Go & ASCII board, SGF and standard coordinate, recent moves, background & temperature 0; thinking disabled & SGF mainline parsing with board rendering before the current move. \\
League of Legends & Video input, match background, API transition context & temperature 0; thinking disabled & Clip-aligned visual input plus team/player state transitions. \\
\bottomrule
\end{tabular}
\caption{Candidate commentary generation settings used by the provided scripts.}
\label{tab:generation-settings}
\end{table*}

The prompts below are copied from the candidate-generation scripts. Runtime placeholders are filled by the corresponding script immediately before the model call.

\section{Evaluation}
\label{app:evaluation}

The evaluation code implements \TACE{} as a namespaced pipeline with four functional branches: segmentation, descriptive evaluation, analytical evaluation, and background evaluation. The run scope can be set to \texttt{all}, \texttt{segmentation}, \texttt{descriptive}, \texttt{analytical}, or \texttt{background}. Judge and extraction calls use temperature 0.0, and sport aliases are normalized before evaluation so that \texttt{football}, \texttt{soccer}, and their Chinese alias map to the same branch.

\subsection{Segmentation for evaluation}

Before type-specific scoring, candidate commentary is segmented into the smallest independent semantic units and assigned category IDs 1, 2, or 3. The implementation validates the returned JSON array and then routes units with ID 1 to descriptive evaluation, ID 2 to analytical evaluation, and ID 3 to background evaluation. The football prompt is shown below as the concrete representative; all other sports use parallel prompt files with the same output contract.

\subsection{Descriptive evaluation}

Descriptive evaluation measures whether the generated commentary correctly captures the salient observable events. The evaluator first obtains a non-empty reference event list from the human reference and state context. It then extracts generated events from candidate descriptive units. For sports, the reference extractor uses labels and match background; for chess and Go, it uses the symbolic board context; for League of Legends, it uses both commentary and API-derived state changes. The same reference event list is used for both completeness and correctness matching. If no descriptive candidate event is extracted, both metrics are set to zero. Scores are rounded to two decimals in the implementation.

\subsection{Analytical evaluation}

Analytical commentary is evaluated through a proposition graph rather than a holistic score. The pipeline first extracts explicit analytical propositions from category-2 units. It then builds sparse forward edge candidates using adjacent propositions and hybrid retrieval, where similarity combines embedding similarity, entity overlap, and keyword overlap. Candidate edges are restricted to the relation set \texttt{CAUSE}, \texttt{REASON}, \texttt{CONDITION}, \texttt{CONTRAST}, and \texttt{ELABORATION}. A second audit stage removes redundant or weakly anchored edges. The resulting graph is validated as a DAG with no self-loop and no duplicate edge for the same source-target pair.

The released implementation stores several analytical diagnostics. The TACE faithfulness signal is the graph-grounded correctness ratio, computed from node and final-edge truth labels:
\begin{equation}
\mathrm{AnalyticalFaithfulness}=\frac{N_{\mathrm{true}}+E_{\mathrm{true}}}{N+E},
\end{equation}
where \(N\) and \(E\) are the number of judged propositions and graph edges. The code also reports \texttt{proposition\_truth}, which is the node-only truth ratio. Reasoning depth is reported through \texttt{effective\_depth}: after node and edge verification, the evaluator keeps only true nodes and true edges, computes the longest valid unweighted DAG path, and normalizes it by the maximum raw depth among models for the same sport-record group. Raw graph \texttt{depth} and \texttt{relation\_info} are retained as additional diagnostics.

\subsection{Background evaluation}
\label{rubric1}

Background evaluation first converts category-3 units into verifiable claims. Each claim carries a source sentence ID, claim type, focus key, entities, and time scope, making the later factuality and relevance judgments operate on independently checkable units.

Factuality is judged against a local background knowledge base derived from the reference record metadata. The retriever combines structured channels based on source ID, focus key, entity alignment, fact-key alignment, and source-context fallback with optional semantic embedding retrieval. The evidence package is bounded by a hard evidence-overflow limit, and the factuality judge may only cite evidence IDs that were actually retrieved. If no valid evidence is available, the claim is labeled false by construction. Relevance is judged separately by comparing each claim with the immediate descriptive and analytical clauses of the same candidate commentary. The reported scores:
\begin{align}
\mathrm{Factuality} &= \frac{\#\text{claims labeled true}}{\#\text{scored claims}},\\
\mathrm{Relevance} &= \frac{\#\text{claims labeled relevant}}{\#\text{scored claims}}.
\end{align}
Both scores are rounded to four decimals; if no claim can be scored, the corresponding value is \texttt{None}.

\section{Human Validation}

\begin{table}[t]
\centering
\caption{Human validation accuracy of LLM-based type-aware decomposition across games.}
\label{tab:type_decomposition_acc}
\small
\setlength{\tabcolsep}{8pt}
\begin{tabular}{lc}
\toprule
\textbf{Game} & \textbf{Accuracy} \\
\midrule
Chess      & 94.0\% \\
Go         & 93.0\% \\
Football   & 98.0\% \\
Basketball & 98.0\% \\
LoL        & 96.0\% \\
\midrule
Overall    & 95.8\% \\
\bottomrule
\end{tabular}
\end{table}

Human annotators were given sampled commentary units or intermediate evaluation items with the corresponding game context when applicable. For type-aware decomposition, they independently assigned each unit one of three labels: descriptive, analytical, or background. For component validation, they independently annotated the target event, matching relation, reasoning relation, verification label, or background-claim judgment according to the provided context and annotation definitions. Their annotations were then compared with the LLM-produced outputs to compute agreement. Human validation was conducted by trained internal annotators following predefined annotation guidelines. No external crowdworkers or paid participant pools were recruited, no monetary compensation was provided. Annotators were instructed to rely only on the provided materials, and no personal information about annotators was collected.

We conduct additional human validation for the LLM-based type-aware decomposition step. For each game, we randomly sample 100 commentary units after decomposition and manually annotate their commentary types as descriptive, analytical, or background. The LLM-produced labels are then compared with the human annotations to compute accuracy.

As shown in Table~\ref{tab:type_decomposition_acc}, the decomposition step achieves consistently high accuracy across all games, ranging from 93.0\% on Go to 98.0\% on football and basketball. The overall accuracy reaches 95.8\%, indicating that the type-aware decomposition used by TACE is reliable across heterogeneous game domains.

\section{Cross-Judge Consistency Check}
\label{cc}

Because TACE uses GPT-5.4-mini for several LLM-mediated evaluation steps, we further examine whether the main analytical results are driven by judge-model preference. This concern is most relevant for analytical commentary: in descriptive evaluation, the key event-matching step is performed by GLM-4.7 rather than a GPT-based judge; in background evaluation, the best-performing models are not GPT-family models. In contrast, analytical node and edge verification are implemented with GPT-5.4-mini, while GPT-5.4 achieves the strongest analytical performance in the main results. We therefore conduct a targeted cross-judge consistency check for analytical metrics.

Specifically, on a held-out subset, we replace the GPT-5.4-mini-based node and edge verification judge with Gemini-3-Flash, while keeping the rest of the TACE pipeline unchanged. We then compare the resulting model rankings with those produced by the main judge using Spearman rank correlation. As shown in Table~\ref{tab:judge_consistency}, the rankings remain consistent, especially for Reasoning Depth. This suggests that the analytical conclusions are not solely driven by preference from the GPT-based judge.

\begin{table}[t]
\centering
\caption{\textbf{Cross-judge consistency for analytical metrics.}
We compare model rankings produced by the main GPT-5.4-mini-based judge and an alternative Gemini-3-Flash-based judge on a held-out subset.}
\label{tab:judge_consistency}
\small
\setlength{\tabcolsep}{5pt}
\begin{tabular}{llc}
\toprule
\textbf{Type} & \textbf{Metric} & \textbf{Spearman $\rho$} \\
\midrule
\multirow{2}{*}{Analytical}
& Faithfulness & 0.77 \\
& Reasoning Depth & 0.94 \\
\bottomrule
\end{tabular}
\end{table}

\onecolumn

\section{Detailed Prompts}

\subsection{Prompts for Data Processing}

\begin{promptbox}{Prompt A.1: Football sentence segmentation and type classification}
You are a professional football commentary analysis expert. Your task is to precisely segment football commentary text into individual sentences and classify each sentence semantically.

\textbf{Football Commentary Classification Standards.}
\begin{enumerate}[leftmargin=*, itemsep=2pt, topsep=2pt]
  \item \textbf{Descriptive Commentary (ID: 1)}
  \begin{itemize}[leftmargin=1.5em, itemsep=1pt, topsep=1pt]
    \item \textbf{Core Question:} ``What is happening on the pitch?''
    \item \textbf{Characteristics:} Objective narration of events and actions directly observable in the game.
    \item \textbf{Includes:} Player movements (shooting, passing, dribbling, defending), score changes, referee calls, possession changes, etc.
    \item \textbf{Examples:} ... ...
  \end{itemize}
  \item \textbf{Analytical Commentary (ID: 2)}
  \begin{itemize}[leftmargin=1.5em, itemsep=1pt, topsep=1pt]
    \item \textbf{Core Question:} ``Why is this happening? What will happen next?''
    \item \textbf{Characteristics:} Explains tactical intentions, evaluates decision quality, predicts game developments, analyzes technical details.
    \item \textbf{Includes:} Tactical analysis, decision evaluation, situation prediction, technical movement breakdown.
    \item \textbf{Examples:} ... ...
  \end{itemize}
  \item \textbf{Background Commentary (ID: 3)}
  \begin{itemize}[leftmargin=1.5em, itemsep=1pt, topsep=1pt]
    \item \textbf{Core Question:} ``What is the background and significance of this event?''
    \item \textbf{Characteristics:} Introduces external knowledge such as player backgrounds, historical data, version information, off-pitch stories, etc.
    \item \textbf{Includes:} Player career history, head-to-head records, injury status, transfer background, statistical data.
    \item \textbf{Examples:} ... ...
  \end{itemize}
\end{enumerate}

\textbf{Important Rules.}
\begin{itemize}[leftmargin=*, itemsep=2pt, topsep=2pt]
  \item Must accurately split the original text into complete sentences; do not merge or over-split.
  \item Each sentence can only be assigned one most appropriate category.
  \item Output must be in strict JSON array format, with each element as a binary tuple \code{[sentence, category_id]}.
  \item Only output the JSON array, with no other text or explanations.
\end{itemize}

\textbf{User Prompt Template.}
\begin{PromptCode}
Please analyze the following football commentary text:

\pvar^commentary~

Perform sentence segmentation and classification according to the above standards, and output the JSON array.
\end{PromptCode}
\end{promptbox}

\begin{promptbox}{Prompt A.2: Basketball sentence segmentation and type classification}
You are a professional basketball commentary analysis expert. Your task is to precisely segment basketball commentary text into individual sentences and classify each sentence semantically.

\textbf{Basketball Commentary Classification Standards.}
\begin{enumerate}[leftmargin=*, itemsep=2pt, topsep=2pt]
  \item \textbf{Descriptive Commentary (ID: 1)}
  \begin{itemize}[leftmargin=1.5em, itemsep=1pt, topsep=1pt]
    \item \textbf{Core Question:} ``What is happening on the court?''
    \item \textbf{Characteristics:} Objective narration of events and actions directly observable in the game.
    \item \textbf{Includes:} Player movements (shooting, driving, passing, defending), score changes, referee calls, possession changes, etc.
    \item \textbf{Examples:} ... ...
  \end{itemize}
  \item \textbf{Analytical Commentary (ID: 2)}
  \begin{itemize}[leftmargin=1.5em, itemsep=1pt, topsep=1pt]
    \item \textbf{Core Question:} ``Why is this happening? What will happen next?''
    \item \textbf{Characteristics:} Explains tactical intentions, evaluates decision quality, predicts game developments, analyzes technical details.
    \item \textbf{Includes:} Tactical analysis, decision evaluation, situation prediction, technical movement breakdown.
    \item \textbf{Examples:} ... ...
  \end{itemize}
  \item \textbf{Background Commentary (ID: 3)}
  \begin{itemize}[leftmargin=1.5em, itemsep=1pt, topsep=1pt]
    \item \textbf{Core Question:} ``What is the background and significance of this event?''
    \item \textbf{Characteristics:} Introduces external knowledge such as player backgrounds, historical data, version information, off-court stories, etc.
    \item \textbf{Includes:} Player career history, head-to-head records, injury status, trade background, statistical data.
    \item \textbf{Examples:} ... ...
  \end{itemize}
\end{enumerate}

\textbf{Important Rules.}
\begin{itemize}[leftmargin=*, itemsep=2pt, topsep=2pt]
  \item Must accurately split the original text into complete sentences; do not merge or over-split.
  \item Each sentence can only be assigned one most appropriate category.
  \item Output must be in strict JSON array format, with each element as a binary tuple \code{[sentence, category_id]}.
  \item Only output the JSON array, with no other text or explanations.
\end{itemize}

\textbf{User Prompt Template.}
\begin{PromptCode}
Please analyze the following basketball commentary text:

\pvar^commentary~

Perform sentence segmentation and classification according to the above standards, and output the JSON array.
\end{PromptCode}
\end{promptbox}

\begin{promptbox}{Prompt A.3: Caption-grounded sentence filtering (football)}
Analyze each sentence independently, applying minimal paraphrasing only to resolve ambiguous references or spelling errors.

\textbf{Retention Criteria} (ALL must be satisfied):
\begin{enumerate}[leftmargin=*, itemsep=2pt, topsep=2pt]
  \item \textbf{Reference Clarity:} Pronouns must clearly refer to a specific entity inferable from caption and \textbf{previously retained} commentary. \textbf{Pronoun inference must be completely unambiguous and free of any doubt.}
  \begin{itemize}[leftmargin=1.5em, itemsep=1pt, topsep=1pt]
    \item \textbf{Mandatory first-sentence truncation check:} Reject if the opening sentence appears incomplete (e.g., begins with conjunction/preposition/subordinate clause or lacks subject/verb). \textbf{In the first sentence, any pronoun referring to a player or team with uncertainty must be rejected, as it likely refers to preceding context that was truncated and cannot be definitively confirmed.}
  \end{itemize}
  \item \textbf{Relevance:} Content must directly relate to caption's action, event, or mentioned players.
  \begin{itemize}[leftmargin=1.5em, itemsep=1pt, topsep=1pt]
    \item \textbf{Category 1 (Action):} Describe caption's action or directly related on-field situations.
    \item \textbf{Category 2 (Strategy):} Analyze tactics based on caption's action.
    \item \textbf{Category 3 (Background):} Provide supplemental info ONLY for players appearing in caption; \textbf{MUST clearly and unambiguously refer to a specific player or team through inference that is completely unambiguous and free of doubt.}
  \end{itemize}
  \item \textbf{Semantic Consistency:} Must not contradict caption or retained sentences.
  \item \textbf{Minimal Correction:} Rephrase solely for:
  \begin{itemize}[leftmargin=1.5em, itemsep=1pt, topsep=1pt]
    \item Ambiguous but inferable references.
    \item Obvious name spelling errors (common in audio transcription).
    \item Preserve original tone and intent.
  \end{itemize}
\end{enumerate}

\textbf{Core Principles:}
\begin{itemize}[leftmargin=*, itemsep=2pt, topsep=2pt]
  \item Caption is authoritative; do not question its accuracy.
  \item \textbf{Only retained sentences may serve as context for subsequent evaluations}.
  \item \textbf{Background information must have a clearly inferable target; reject if the player/team reference is ambiguous or uncertain}.
  \item Violation of criteria 1--3 $\rightarrow$ reject without paraphrasing.
\end{itemize}

\textbf{Output Format.}

\textbf{Part 1: Analysis.} For each sentence, detail:
\begin{itemize}[leftmargin=*, itemsep=2pt, topsep=2pt]
  \item Reference clarity (with truncation assessment and \textbf{pronoun ambiguity evaluation, especially for player/team references in first sentence}).
  \item Relevance and factual accuracy (state reason if failed; for Category 3, \textbf{explicitly flag target specificity and inference clarity concerns}).
  \item Semantic consistency (identify contradictions if failed).
  \item \textbf{Reference Specificity (Category 3 ONLY):} Confirm background information clearly and unambiguously refers to a specific player or team through inference that is completely unambiguous and free of doubt; reject if the target is unclear or requires ambiguous inference.
  \item Paraphrasing need (specify changes and rationale).
  \item Final verdict (retain/reject; provide corrected version if applicable).
\end{itemize}

\textbf{Part 2: JSON.}
\begin{PromptCode}
[
  ["Sentence1", Category1, true/false],
  ["Sentence2", Category2, true/false],
  ...
]
\end{PromptCode}
\begin{itemize}[leftmargin=*, itemsep=2pt, topsep=2pt]
  \item Strict array of triplets: \code{[sentence, category, retention flag]}.
  \item Sentence: original if unchanged, corrected if paraphrased.
  \item Retention flag: lowercase \code{true/false}.
  \item Category number unchanged.
\end{itemize}

\textbf{User Prompt Template.}
\begin{PromptCode}
Caption (authoritative action description):
\pvar^caption~

Commentary Sentences (pre-classified):
\pvar^sentences_info~

Apply the specified retention criteria to evaluate each sentence independently. Output Part 1: Analysis and Part 2: JSON.
\end{PromptCode}
\end{promptbox}


\begin{promptbox}{Prompt A.4: Football caption-commentary fusion}
You are an expert soccer video content analyst. Your task is to combine the caption and commentary into a coherent narrative written in the voice of a professional sports commentator.

\textbf{Input:}
\begin{itemize}[leftmargin=*, itemsep=2pt, topsep=2pt]
  \item Caption: A brief textual description of a soccer/football game clip, focusing on the main events and player actions occurring in the video.
  \item Commentary: Transcribed live commentary from the same video segment, which may contain multiple pieces of information.
\end{itemize}

\textbf{Task:}
Combine the caption and commentary into a coherent, natural-sounding description that captures the key elements of both, written \textbf{in the live broadcast voice while maintaining stylistic fidelity to the provided commentary---mirroring its energy, pacing, vocabulary, and tonal qualities}. The result should read like a unified live commentary of the event, not as two separate pieces of information.

\textbf{Critical Filtering Rule:}
Before combining, carefully evaluate each piece of information in the commentary according to these strict hierarchical requirements:
\begin{enumerate}[leftmargin=*, itemsep=2pt, topsep=2pt]
  \item \textbf{Event Description Criterion:} Commentary elements describing on-field actions must demonstrate direct relevance to the caption's events OR establish a clear chronological relationship with them. Exclude isolated event descriptions that lack temporal or causal connection to the caption narrative.
  \item \textbf{Tactical Analysis Criterion:} Strategic or tactical observations must be explicitly anchored to specific event descriptions from the caption or retained commentary. Exclude standalone tactical analysis that references no concrete on-field action present in the source materials.
  \item \textbf{Background Information Criterion:} Contextual details (player positioning, formations, situational factors) may only be included when they directly supplement and illuminate events described in the caption or retained commentary. Exclude background information wholly detached from the depicted action.
  \item \textbf{General Enhancement Criterion:} Only retain commentary elements that satisfy the above three criteria and demonstrably enhance or complement the caption's narrative while maintaining absolute factual consistency.
\end{enumerate}

\textbf{Strict Factual Boundaries:}
Your entire narrative must be constructed \textbf{exclusively} from information explicitly present in the caption and commentary. You are prohibited from:
\begin{itemize}[leftmargin=*, itemsep=2pt, topsep=2pt]
  \item Introducing any external facts, statistics, or context not provided in the source materials.
  \item Speculating on player motivations, emotions, or unspoken thoughts.
  \item Fabricating details about game situation (score, time, half) beyond stated information.
  \item Assuming or inventing player names, teams, venues, or historical data.
  \item Creating hypothetical scenarios or ``what-if'' scenarios.
\end{itemize}

\textbf{Focus on:}
\begin{enumerate}[leftmargin=*, itemsep=2pt, topsep=2pt]
  \item Creating a seamless narrative \textbf{that authentically reflects the source commentary's unique style, delivery patterns, and rhetorical approach}, incorporating both factual events (caption) and relevant descriptive/color commentary.
  \item Maintaining strict chronological flow of events as presented in the caption.
  \item \textbf{Employing the specific language patterns, excitement cadence, and terminology preferences exhibited in the provided commentary}, ensuring stylistic consistency while applying professional broadcasting standards.
  \item Ensuring clarity, readability, and factual accuracy by filtering out irrelevant or contradictory commentary.
  \item Preserving all important information from the caption while selectively incorporating valuable commentary elements that enhance the broadcast-style narrative.
  \item Maintaining absolute fidelity to the source material, with zero speculation or fabrication.
\end{enumerate}

\textbf{Output Format:}
Present your combined narrative inside a box using the format \verb|\boxed{your_combined_narrative_here}|. This should be the only content in your response. The narrative must be written \textbf{in the distinctive style and tone captured in the source commentary}.

\textbf{User Prompt Template.}
\begin{PromptCode}
Caption: \pvar^caption~
Commentary:
\pvar^commentary_text~
\end{PromptCode}
\end{promptbox}

\begin{promptbox}{Prompt A.5: Basketball caption-commentary relevance gate}
You are an expert basketball video content analyst. Your task is to determine whether a live commentary segment is semantically relevant to the current video content described in a caption.

\textbf{Input:}
\begin{itemize}[leftmargin=*, itemsep=2pt, topsep=2pt]
  \item Caption: A brief textual description of a basketball game clip, focusing on the main events and player actions occurring in the video (e.g., ``Player X shoots a three-pointer'', ``Team Y executes a fast break'').
  \item Commentary: Transcribed live commentary from the same video segment, which may have temporal lag and could discuss events beyond the immediate clip.
\end{itemize}

\textbf{Relevance Criteria:}
The commentary is considered RELEVANT (true) if:
\begin{enumerate}[leftmargin=*, itemsep=2pt, topsep=2pt]
  \item It directly describes or comments on the events/actions mentioned in the caption, OR
  \item It provides strategic analysis of team tactics based on the caption's events, OR
  \item It offers background context about players/teams mentioned in the caption (e.g., season statistics, career achievements, team history), OR
  \item It discusses game implications (score changes, momentum shifts) related to the caption's events.
\end{enumerate}
The commentary is considered IRRELEVANT (false) if:
\begin{enumerate}[leftmargin=*, itemsep=2pt, topsep=2pt]
  \item It discusses completely unrelated events (different plays, previous games, off-court topics), OR
  \item It contains generic filler content with no connection to the caption's specific events or entities.
\end{enumerate}

\textbf{Output Format.}
\begin{itemize}[leftmargin=*, itemsep=2pt, topsep=2pt]
  \item \textbf{Phase 1 (Analysis):} Provide a concise reasoning analyzing the semantic relationship between caption and commentary. Consider temporal context, entity mentions, and thematic connections.
  \item \textbf{Phase 2 (Judgment):} Output ONLY the final judgment as \verb|\boxed{true}| or \verb|\boxed{false}| with no additional text.
\end{itemize}

\textbf{Important:} Your analysis must be objective and based solely on semantic relevance. Do not over-interpret vague connections. The boxed judgment must be your definitive conclusion.

\textbf{User Prompt Template.}
\begin{PromptCode}
caption: \pvar^caption~
commentary: \pvar^commentary~
\end{PromptCode}
\end{promptbox}

\begin{promptbox}{Prompt A.6: Basketball caption-commentary fusion}
You are an expert basketball video content analyst. Your task is to combine the caption and commentary into a coherent narrative written in the voice of a professional sports commentator.

\textbf{Input:}
\begin{itemize}[leftmargin=*, itemsep=2pt, topsep=2pt]
  \item Caption: A brief textual description of a basketball game clip, focusing on the main events and player actions occurring in the video.
  \item Commentary: Transcribed live commentary from the same video segment, which may contain multiple pieces of information.
\end{itemize}

\textbf{Task:}
Combine the caption and commentary into a coherent, natural-sounding description that captures the key elements of both, written \textbf{in the live broadcast voice while maintaining stylistic fidelity to the provided commentary---mirroring its energy, pacing, vocabulary, and tonal qualities}. The result should read like a unified live commentary of the event, not as two separate pieces of information.

\textbf{Critical Filtering Rule:}
Before combining, carefully evaluate each piece of information in the commentary:
\begin{itemize}[leftmargin=*, itemsep=2pt, topsep=2pt]
  \item If a commentary element is directly supported by or consistent with the caption's events, include it.
  \item If a commentary element is unrelated to the caption's described events, exclude it.
  \item If a commentary element contradicts the factual events described in the caption, exclude it.
  \item Only retain commentary elements that enhance or complement the caption's narrative.
\end{itemize}

\textbf{Strict Factual Boundaries:}
Your entire narrative must be constructed \textbf{exclusively} from information explicitly present in the caption and commentary. You are prohibited from:
\begin{itemize}[leftmargin=*, itemsep=2pt, topsep=2pt]
  \item Introducing any external facts, statistics, or context not provided in the source materials.
  \item Speculating on player motivations, emotions, or unspoken thoughts.
  \item Fabricating details about game situation (score, time, quarter) beyond stated information.
  \item Assuming or inventing player names, teams, venues, or historical data.
  \item Creating hypothetical scenarios or ``what-if'' scenarios.
\end{itemize}

\textbf{Focus on:}
\begin{enumerate}[leftmargin=*, itemsep=2pt, topsep=2pt]
  \item Creating a seamless narrative \textbf{that authentically reflects the source commentary's unique style, delivery patterns, and rhetorical approach}, incorporating both factual events (caption) and relevant descriptive/color commentary.
  \item Maintaining strict chronological flow of events as presented in the caption.
  \item \textbf{Employing the specific language patterns, excitement cadence, and terminology preferences exhibited in the provided commentary}, ensuring stylistic consistency while applying professional broadcasting standards.
  \item Ensuring clarity, readability, and factual accuracy by filtering out irrelevant or contradictory commentary.
  \item Preserving all important information from the caption while selectively incorporating valuable commentary elements that enhance the broadcast-style narrative.
  \item Maintaining absolute fidelity to the source material, with zero speculation or fabrication.
\end{enumerate}

\textbf{Output Format:}
Present your combined narrative inside a box using the format \verb|\boxed{your_combined_narrative_here}|. This should be the only content in your response. The narrative must be written \textbf{in the distinctive style and tone captured in the source commentary}.

\textbf{User Prompt Template.}
\begin{PromptCode}
Caption: \pvar^caption~
Commentary: \pvar^commentary~
\end{PromptCode}
\end{promptbox}

\begin{promptbox}{Prompt A.7: Basketball quality selection prompt}
\textbf{System Prompt.}

You are a professional sports-commentary analyst. From a batch of basketball game commentaries, select the two highest-quality items.

\textbf{Evaluation Criteria:} Each commentary will be scored on a 10-point scale across the following three dimensions, and the final ranking is based on the total score:
\begin{enumerate}[leftmargin=*, itemsep=2pt, topsep=2pt]
  \item \textbf{Descriptive Dimension (Descriptive):} Evaluate the precision and information density with which the commentary captures the real-time game picture: whether observable facts such as the ball-handler, tactical movement, screen quality, shot area, defensive matchup, and referee calls are stated clearly, completely, and in detail. A higher score means the account of ``what is happening at this moment'' is more like a precise courtside documentary.
  \item \textbf{Analytical Dimension (Analytical):} Evaluate the depth of interpretation of the game's underlying logic: whether offensive actions (Spain P\&R, Horns, Flare), defensive strategies (switching, double-teaming, ICE), player decision quality, coaching matchups, and tactical trade-offs under score/time/foul situations are analyzed thoroughly. A higher score means the insight into ``why the play unfolds this way'' is more like that of a tactical master.
  \item \textbf{Background Dimension (Background):} Evaluate the ability to extend the event context: whether player historical statistics, matchup histories, team season trends, key-game conventions, cultural references (e.g., ``Christmas Day game''), sneakers, records, and other off-court information are naturally integrated to add the ``deeper meaning of this moment'' to the current picture. A higher score means the rendering of ``what this moment means'' is more like that of a storyteller.
\end{enumerate}

\textbf{Analysis Requirements:} Follow these steps:
\begin{itemize}[leftmargin=*, itemsep=2pt, topsep=2pt]
  \item \textbf{Step 1 (Analysis Phase):} For each commentary, provide analysis and a score for each of the three dimensions on a 10-point scale, then calculate the total score.
  \item \textbf{Step 2 (Selection Phase):} Sort the commentaries by total score from high to low, select the two highest-scoring commentaries, and return only their \code{video_id} values.
\end{itemize}

\textbf{Output Format:}
\begin{PromptCode}
[Analysis Phase]
1. [video_id]:
   - Descriptive Dimension (Descriptive): [Reason] - X points
   - Analytical Dimension (Analytical): [Reason] - X points
   - Background Dimension (Background): [Reason] - X points
   - Total Score: XX points

...

[Selection Phase]
["video_id1", "video_id2"]
\end{PromptCode}

\textbf{User Prompt Template.}
\begin{PromptCode}
Select the two highest-quality commentaries from the following 3 basketball-game commentaries:

\pvar^commentary_list~
\end{PromptCode}
\end{promptbox}

\subsection{Prompts for Commentary Generation}

\begin{promptbox}{Prompt B.1: Football / Basketball candidate-generation prompt}
\textbf{Role:}
You are a top-tier professional \blue{football} / \red{basketball} commentator. Your task is to generate an immersive, high-quality live commentary track based on the provided description of a match video clip.

\textbf{Core Objective:}
Produce a coherent, pure-text narrative that replicates the energy, pacing, and terminology of a live TV broadcast. The output must be spoken directly to the audience (first-person active voice), avoiding third-party summarization styles.

\textbf{Content Requirements (Semantic Layers):}
Your commentary must align with the three core commentator capabilities and their corresponding commentary types:
\begin{enumerate}[leftmargin=*, itemsep=2pt, topsep=2pt]
  \item \textbf{Live Observation $\rightarrow$ Descriptive Commentary:}
  \begin{itemize}[leftmargin=1.5em, itemsep=1pt, topsep=1pt]
    \item Accurately perceive and interpret real-time events and fine-grained details in the game environment.
    \item Objectively answer ``What is happening?'' by narrating immediate, observable events.
  \end{itemize}
  \item \textbf{Strategic Analysis $\rightarrow$ Analytical Commentary:}
  \begin{itemize}[leftmargin=1.5em, itemsep=1pt, topsep=1pt]
    \item Comprehend the current game state, interpret strategic intent, evaluate decisions, and anticipate future developments.
    \item Go beyond description by assessing ``How good?'', explaining ``Why?'', and predicting ``What's next?'' within a broader tactical context.
  \end{itemize}
  \item \textbf{Historical Recall $\rightarrow$ Background Commentary:}
  \begin{itemize}[leftmargin=1.5em, itemsep=1pt, topsep=1pt]
    \item Leverage extensive game knowledge, including player histories, classic match moments, and community-specific terminology.
    \item Answer ``What is the significance?'' by integrating broader contextual information such as detailed team and player backgrounds, game history, and cultural references.
  \end{itemize}
\end{enumerate}

\textbf{Style \& Tone Guidelines:}
\begin{itemize}[leftmargin=*, itemsep=2pt, topsep=2pt]
  \item \textbf{Dynamic Pacing:} Match the rhythm of the text to the intensity of the event (e.g., short, punchy sentences for goals; longer, analytical flows for build-up play).
  \item \textbf{Professional Vocabulary:} Use authentic \blue{football} / \red{basketball} terminology.
  \item \textbf{Natural Fluency:} Avoid robotic sentence structures. Use varied syntax and emotive language characteristic of human commentators.
\end{itemize}

\textbf{Strict Constraints:}
\begin{enumerate}[leftmargin=*, itemsep=2pt, topsep=2pt]
  \item \textbf{Format:} Output raw text only as a \textbf{single continuous paragraph}. \textbf{Do not use line breaks} (\texttt{\textbackslash n}), Markdown, bolding, labels, or list formats.
  \item \textbf{Language:} Maintain a single language consistently. Use \textbf{English} uniformly.
  \item \textbf{Factual Grounding:} Do not hallucinate scorelines, team names, or events not implied by the input. Do not invent specific statistics unless provided.
  \item \textbf{Consistency:} Ensure the narrative flow is logical and non-contradictory.
  \item \textbf{Scope:} The video is only a segment of the match. The commentary must focus exclusively on this segment, not the full game.
\end{enumerate}

\textbf{User Prompt Template.}
\begin{PromptCode}
Match Background (Full Match Context, not limited to this clip):
\pvar^background~

Generate a \blue^football~ / \red^basketball~ commentary based on the video content. The video segment is \pvar^duration_seconds~ seconds long. Complete the entire commentary within \pvar^duration_seconds~ seconds.
\end{PromptCode}
\end{promptbox}

\begin{promptbox}{Prompt B.2: Chess / Go candidate-generation prompt}
\textbf{System Prompt.}
You are a top-tier professional \blue{chess} / \red{go} commentator. Your task is to generate an immersive, high-quality move commentary based on the provided game state information.

\textbf{User Prompt Template.}

\textbf{Role:}
You are a top-tier professional \blue{chess} / \red{go} commentator. Your task is to generate an immersive, high-quality move-by-move commentary based on the provided game state.

\textbf{Core Objective:}
Produce a coherent, pure-text narrative that replicates the insight, pacing, and terminology of a professional \blue{chess} / \red{go} broadcast. The output must be spoken directly to the audience (first-person active voice), avoiding third-party summarization styles.

\textbf{Content Guidelines (Semantic Layers):}
You may draw on the following three core commentary capabilities of good commentators. These are provided as flexible guidance rather than requirements, and may be blended naturally.
\begin{enumerate}[leftmargin=*, itemsep=2pt, topsep=2pt]
  \item \textbf{Move Observation $\rightarrow$ Descriptive Commentary:}
  \begin{itemize}[leftmargin=1.5em, itemsep=1pt, topsep=1pt]
    \item Accurately perceive and describe the current move, the piece involved, and the squares it moves from and to.
    \item Objectively answer ``What is happening on the board?'' by narrating the immediate, observable move.
  \end{itemize}
  \item \textbf{Strategic Analysis $\rightarrow$ Analytical Commentary:}
  \begin{itemize}[leftmargin=1.5em, itemsep=1pt, topsep=1pt]
    \item Comprehend the current board position (provided via FEN), interpret the tactical or positional intent behind the move, evaluate its quality, and anticipate future developments.
    \item Go beyond description by assessing ``How good is this move?'', explaining ``Why was it played?'', and predicting ``What's next?'' within a broader strategic context.
  \end{itemize}
  \item \textbf{Historical Recall $\rightarrow$ Background Commentary:}
  \begin{itemize}[leftmargin=1.5em, itemsep=1pt, topsep=1pt]
    \item Leverage extensive \blue{chess} / \red{go} knowledge, including opening theory, classic game references, and player tendencies.
    \item Answer ``What is the significance?'' by integrating broader contextual information such as player backgrounds, tournament stakes, and historical parallels.
  \end{itemize}
\end{enumerate}

\textbf{Style \& Tone Guidelines:}
\begin{itemize}[leftmargin=*, itemsep=2pt, topsep=2pt]
  \item \textbf{Professional Vocabulary:} Use authentic \blue{chess} / \red{go} terminology.
  \item \textbf{Natural Fluency:} Avoid robotic sentence structures. Use varied syntax and emotive language characteristic of human commentators.
\end{itemize}

\textbf{Strict Constraints:}
\begin{enumerate}[leftmargin=*, itemsep=2pt, topsep=2pt]
  \item \textbf{Format:} Output raw text only as a \textbf{single continuous paragraph}. \textbf{Do not use line breaks} (\texttt{\textbackslash n}), Markdown, bolding, labels, or list formats.
  \item \textbf{Language:} Use \textbf{English} uniformly.
  \item \textbf{Factual Grounding:} Do not hallucinate player names, ratings, or events not implied by the input. Do not invent specific statistics unless provided.
  \item \textbf{Consistency:} Ensure the narrative flow is logical and non-contradictory.
  \item \textbf{Scope:} The commentary can focus on this specific move and its immediate context or the entire game.
\end{enumerate}

\begin{PromptCode}
Match Background (Full Game Context):
\pvar^background~

\blue^Game State:
Board before current move (FEN): \pvar^fen_before~
Current move: \pvar^full_move_str~ (SAN=\pvar^san~, UCI=\pvar^uci~)
Board after current move (FEN): \pvar^fen_after~
Previous moves (up to last 3 full moves by both sides): \pvar^history_text~~

\red^
Game State:
Board before current move (X=Black, O=White, .=empty):
\pvar^board_ascii~

Current move (ply \pvar^ply~): \pvar^move_str~ (SGF coord=\pvar^coord_sgf~, standard=\pvar^coord_std~)
Recent moves (last 3 full rounds): \pvar^history_text~~

Generate a \blue^chess~ / \red^go~ commentary for this move.
\end{PromptCode}
\end{promptbox}

\begin{promptbox}{Prompt B.3: League of Legends candidate-generation prompt}
\textbf{System Prompt.}
You are a top-tier professional League of Legends commentator. Your task is to generate an immersive, high-quality live commentary track based on the provided match video clip.

\textbf{User Prompt Template.}

\textbf{Role:}
You are a top-tier professional League of Legends commentator. Your task is to generate an immersive, high-quality live commentary track based on the provided match video clip.

\textbf{Core Objective:}
Produce a coherent, pure-text narrative that replicates the energy, pacing, and terminology of a live esports broadcast. The output must be spoken directly to the audience (first-person active voice), avoiding third-party summarization styles.

\textbf{Content Guidelines (Semantic Layers):}
You may draw on the following three core commentary capabilities of good commentators. These are provided as flexible guidance rather than requirements, and may be blended naturally.
\begin{enumerate}[leftmargin=*, itemsep=2pt, topsep=2pt]
  \item \textbf{Live Observation $\rightarrow$ Descriptive Commentary:}
  \begin{itemize}[leftmargin=1.5em, itemsep=1pt, topsep=1pt]
    \item Accurately perceive and interpret real-time events and fine-grained details in the game environment.
    \item Objectively answer ``What is happening?'' by narrating immediate, observable events (champion movements, ability usage, objective takes, kills, map states).
  \end{itemize}
  \item \textbf{Strategic Analysis $\rightarrow$ Analytical Commentary:}
  \begin{itemize}[leftmargin=1.5em, itemsep=1pt, topsep=1pt]
    \item Comprehend the current game state, interpret strategic intent, evaluate decisions, and anticipate future developments.
    \item Go beyond description by assessing ``How good?'', explaining ``Why?'', and predicting ``What's next?'' within a broader tactical context (draft implications, win conditions, power spikes, macro decisions).
  \end{itemize}
  \item \textbf{Historical Recall $\rightarrow$ Background Commentary:}
  \begin{itemize}[leftmargin=1.5em, itemsep=1pt, topsep=1pt]
    \item Leverage extensive League of Legends knowledge, including player histories, classic match moments, and community-specific terminology.
    \item Answer ``What is the significance?'' by integrating broader contextual information such as team and player backgrounds, game history, and cultural references.
  \end{itemize}
\end{enumerate}

\textbf{Style \& Tone Guidelines:}
\begin{itemize}[leftmargin=*, itemsep=2pt, topsep=2pt]
  \item \textbf{Professional Vocabulary:} Use authentic League of Legends terminology (e.g., ``lane pressure,'' ``reset,'' ``trade,'' ``skirmish,'' ``rotation,'' ``vision control,'' ``engage,'' ``disengage,'' ``objective posture,'' ``power spike,'' ``team comp'').
  \item \textbf{Natural Fluency:} Avoid robotic sentence structures. Use varied syntax and emotive language characteristic of human casters.
\end{itemize}

\textbf{Strict Constraints:}
\begin{enumerate}[leftmargin=*, itemsep=2pt, topsep=2pt]
  \item \textbf{Format:} Output raw text only as a \textbf{single continuous paragraph}. \textbf{Do not use line breaks} (\texttt{\textbackslash n}), Markdown, bolding, labels, or list formats.
  \item \textbf{Language:} Use \textbf{English} uniformly.
  \item \textbf{Factual Grounding:} Do not hallucinate champion names, gold values, objectives, kills, or map states that are not clearly supported by the video or provided context.
  \item \textbf{Consistency:} Ensure the narrative flow is logical and non-contradictory.
  \item \textbf{Scope:} The video is only a segment of the match. The commentary must focus exclusively on this segment, not the full game.
\end{enumerate}

\begin{PromptCode}
Match Background (Full Match Context, not limited to this clip):
\pvar^background~

\pvar^extra_context_block~

The video segment is \pvar^duration_seconds~ seconds long. Complete the entire commentary within \pvar^duration_seconds~ seconds.

Generate a League of Legends commentary based on the video content.
\end{PromptCode}
\end{promptbox}

\subsection{Prompt for TACE}

\subsubsection{Prompt for Generated Commentary Segmentation}

\begin{promptbox}{Prompt C.1: Evaluation segmentation and type classification (football)}
\textbf{System Prompt.}
You are a professional football commentary analysis expert. Your task is to precisely segment football commentary text into individual sentences and classify each sentence semantically.

\textbf{Football Commentary Classification Standards.}
\begin{enumerate}[leftmargin=*, itemsep=2pt, topsep=2pt]
  \item \textbf{Descriptive Commentary (ID: 1)}
  \begin{itemize}[leftmargin=1.5em, itemsep=1pt, topsep=1pt]
    \item \textbf{Core Question:} ``What is happening on the pitch?''
    \item \textbf{Characteristics:} Objective narration of events and actions directly observable in the game, without subjective evaluation.
    \item \textbf{Includes:} Player movements (shooting, passing, dribbling, defending), score changes, referee calls, possession changes, etc.
    \item \textbf{Examples:} ... ...
  \end{itemize}
  \item \textbf{Analytical Commentary (ID: 2)}
  \begin{itemize}[leftmargin=1.5em, itemsep=1pt, topsep=1pt]
    \item \textbf{Core Question:} ``How good? Why is this happening? What will happen next?''
    \item \textbf{Characteristics:} Subjectively evaluates performance or decisions, explains tactical intentions and causes, predicts game developments, analyzes technical details.
    \item \textbf{Includes:} Tactical analysis, decision evaluation, situation prediction, technical movement breakdown, emotional evaluation of plays.
    \item \textbf{Examples:} ... ...
  \end{itemize}
  \item \textbf{Background Commentary (ID: 3)}
  \begin{itemize}[leftmargin=1.5em, itemsep=1pt, topsep=1pt]
    \item \textbf{Core Question:} ``What is the background and significance of this event?''
    \item \textbf{Characteristics:} Introduces external knowledge such as player backgrounds, historical data, off-pitch stories, etc.
    \item \textbf{Includes:} Player career history, head-to-head records, injury status, transfer background, statistical data.
    \item \textbf{Examples:} ... ...
  \end{itemize}
\end{enumerate}

\textbf{Boundary Rules:}
\begin{itemize}[leftmargin=*, itemsep=2pt, topsep=2pt]
  \item When a sentence contains elements of multiple categories, classify based on its primary intent (semantic focus).
  \item Pure emotional exclamations with evaluative tone (e.g., ``What a goal!'') belong to Analytical Commentary (ID: 2).
  \item Objective data statements (e.g., ``This is his 30th goal'') belong to Background Commentary (ID: 3); interpretation or inference drawn from data belongs to Analytical Commentary (ID: 2).
\end{itemize}

\textbf{Important Rules:}
\begin{itemize}[leftmargin=*, itemsep=2pt, topsep=2pt]
  \item Segment text into the smallest independent semantic units (clauses or short phrases) rather than full grammatical sentences. Split compound sentences at commas or conjunctions if they contain distinct events or judgments.
  \item Each sentence can only be assigned one most appropriate category.
  \item Output must be in strict JSON array format, with each element as a binary tuple \code{[sentence, category_id]}.
  \item Only output the JSON array, with no other text or explanations.
\end{itemize}

\textbf{User Prompt Template.}
\begin{PromptCode}
Please analyze the following football commentary text:

\pvar^commentary~

Perform sentence segmentation and classification according to the above standards, and output the JSON array.
\end{PromptCode}
\end{promptbox}

\subsubsection{Prompt for Descriptive Evaluation}

\begin{promptbox}{Prompt C.2: Reference event extraction (football)}
\textbf{System Prompt.}
You are a football event extraction expert. Your task is to extract the main events from the given commentary and label information.

\textbf{Input:}
\begin{itemize}[leftmargin=*, itemsep=2pt, topsep=2pt]
  \item \code{label}: The main event type of the video clip (e.g., goal, corner, clearance, etc.).
  \item \code{processed_commentary}: The processed reference commentary describing what happened.
  \item \code{background_xml}: Full XML-formatted match background, including teams and player rosters.
\end{itemize}

\textbf{Task:}
Extract the main events that occurred in the video clip. Each event should be described in a single sentence including:
\begin{itemize}[leftmargin=*, itemsep=2pt, topsep=2pt]
  \item Who (player name or team).
  \item What (action).
  \item Result (if applicable).
\end{itemize}

\textbf{Output Format.}
\begin{PromptCode}
["Event 1 description", "Event 2 description", ...]
\end{PromptCode}

\textbf{Important:}
\begin{itemize}[leftmargin=*, itemsep=2pt, topsep=2pt]
  \item Only output the JSON array.
  \item All event descriptions must be in English.
  \item Focus on the main events mentioned in the label and commentary.
  \item Do not include minor details or repetitive descriptions.
\end{itemize}

\textbf{Examples.}
... ...


\textbf{User Prompt Template.}
\begin{PromptCode}
Label: \pvar^label~
Processed Commentary: \pvar^processed_commentary~
Background XML:
\pvar^background~

Extract the main events as a JSON array of strings.
\end{PromptCode}
\end{promptbox}
\begin{promptbox}{Prompt C.3: Candidate event extraction (football)}
\textbf{System Prompt.}
You are a football event extraction expert. Your task is to extract the main events from the given commentary text.

\textbf{Task:}
Extract the main events that occurred in the video clip based on the commentary. Each event should be described in a single sentence including:
\begin{itemize}[leftmargin=*, itemsep=2pt, topsep=2pt]
  \item Who (player name or team).
  \item What (action).
  \item Result (if applicable).
\end{itemize}

\textbf{Output Format.}
\begin{PromptCode}
["Event 1 description", "Event 2 description", ...]
\end{PromptCode}

\textbf{Important:}
\begin{itemize}[leftmargin=*, itemsep=2pt, topsep=2pt]
  \item Only output the JSON array.
  \item All event descriptions must be in English.
  \item Focus on the main events mentioned in the commentary.
  \item Do not include minor details or repetitive descriptions.
\end{itemize}

\textbf{Examples.}
... ...


\textbf{User Prompt Template.}
\begin{PromptCode}
Commentary: \pvar^commentary~

Extract the main events as a JSON array of strings.
\end{PromptCode}
\end{promptbox}
\begin{promptbox}{Prompt C.4: Completeness matching}
\textbf{System Prompt.}
You are a semantic matching expert for completeness evaluation. Your task is to match ground truth events to evaluated events.

\textbf{Input:}
\begin{itemize}[leftmargin=*, itemsep=2pt, topsep=2pt]
  \item Ground Truth Events: The reference events that should be covered (from human-annotated data, high certainty).
  \item Evaluated Events: The events extracted from the commentary being evaluated.
\end{itemize}

\textbf{Task:}
For each event in Ground Truth Events, determine if there is a matching event in Evaluated Events.

\textbf{Strict Matching Criteria:}
\begin{itemize}[leftmargin=*, itemsep=2pt, topsep=2pt]
  \item The action must be highly consistent.
  \item The subject (player/team) must be clearly identified and match.
  \item The result must match (if applicable).
  \item Only match if you are confident they describe the same event.
\end{itemize}

\textbf{Output Format.}
\begin{PromptCode}
{
  "matches": [evaluated_index_or_null, ...],
  "reasoning": "detailed explanation of matching decisions"
}
\end{PromptCode}

\textbf{matches array:}
\begin{itemize}[leftmargin=*, itemsep=2pt, topsep=2pt]
  \item Length must equal Ground Truth Events length.
  \item Each element is the index (0-based) of the matching event in Evaluated Events, or \code{null} if no match.
\end{itemize}

\textbf{Examples.}
... ...


\textbf{User Prompt Template.}
\begin{PromptCode}
Ground Truth Events: \pvar^ground_truth_events~
Evaluated Events: \pvar^evaluated_events~

For each ground truth event, find the matching evaluated event index (or null). Return as JSON with "matches" array and "reasoning".
\end{PromptCode}
\end{promptbox}
\begin{promptbox}{Prompt C.5: Correctness matching}
\textbf{System Prompt.}
You are a semantic matching expert for precision evaluation. Your task is to match evaluated events to ground truth events.

\textbf{Input:}
\begin{itemize}[leftmargin=*, itemsep=2pt, topsep=2pt]
  \item Evaluated Events: The events extracted from the commentary being evaluated.
  \item Ground Truth Events: The reference-text-derived event list, which is considered accurate and should be treated as the evidence source.
\end{itemize}

\textbf{Task:}
For each event in Evaluated Events, determine if there is a matching event in Ground Truth Events.

\textbf{Moderately Strict, Evidence-Conservative Matching Criteria:}
\begin{itemize}[leftmargin=*, itemsep=2pt, topsep=2pt]
  \item The core action must match. Do not match different event categories just because they happen in a similar scene.
  \item The result or outcome must also match when applicable (for example: score vs miss, save vs goal, foul vs clean tackle).
  \item The subject, team, and attacking side must not conflict.
  \item Flexible references are allowed when they can plausibly refer to the same actor without conflict, such as names, jersey numbers, roles, or descriptive references.
  \item Generalization is allowed if the evaluated event is less specific but still supported by the ground truth event.
  \item Do not match if the evaluated event adds unsupported critical details that are not grounded in the reference event, such as a specific player, team, attacking side, event type, decisive result, or key execution detail.
  \item Be conservative: if a critical fact is unsupported or conflicts with the evidence, return \code{null}.
\end{itemize}

\textbf{Output Format.}
\begin{PromptCode}
{
  "matches": [gt_index_or_null, ...],
  "reasoning": "detailed explanation of matching decisions"
}
\end{PromptCode}

\textbf{matches array:}
\begin{itemize}[leftmargin=*, itemsep=2pt, topsep=2pt]
  \item Length must equal Evaluated Events length.
  \item Each element is the index (0-based) of the matching event in Ground Truth Events, or \code{null} if no match.
\end{itemize}

\textbf{Examples.}
... ...


\textbf{User Prompt Template.}
\begin{PromptCode}
Evaluated Events: \pvar^evaluated_events~
Ground Truth Events: \pvar^ground_truth_events~

For each evaluated event, find the matching ground truth event index (or null). Return as JSON with "matches" array and "reasoning".
\end{PromptCode}
\end{promptbox}

\subsubsection{Prompt for Analytical Evaluation}

\begin{promptbox}{Prompt C.6: Analytical relation standard}
\begin{itemize}[leftmargin=*, itemsep=2pt, topsep=2pt]
  \item \textbf{CAUSE:} A directly brings about, triggers, or produces B.
  \item \textbf{REASON:} B provides the cause, explanation, evidence, or justification for A.
  \item \textbf{CONDITION:} A states a hypothetical or unrealized premise; B holds only under that premise.
  \item \textbf{CONTRAST:} A establishes an expectation or direction; B violates or reverses it.
  \item \textbf{ELABORATION:} B is an intrinsic structural component or unpacking of A.
\end{itemize}
\end{promptbox}
\begin{promptbox}{Prompt C.7: Analytical proposition extraction}
\textbf{System Prompt.}
\begin{PromptCode}

\end{PromptCode}

\textbf{User Prompt.}
You extract analytical propositions from stage-1 analytical sentences.

\textbf{Task:}
\begin{itemize}[leftmargin=*, itemsep=2pt, topsep=2pt]
  \item Use the full commentary only as context for reference resolution.
  \item Extract propositions ONLY from the provided analytical sentence records.
  \item Each proposition must remain inside its own source sentence.
  \item Never fuse content across two different analytical sentences.
  \item Preserve sentence order.
  \item Preserve proposition order inside each sentence.
  \item Prefer complete proposition units, not fragments.
\end{itemize}

\textbf{Input Template.}
\begin{PromptCode}
Input commentary:
\pvar^commentary~

Analytical sentence records:
\pvar^analytical_sentence_records_json~
\end{PromptCode}

\textbf{Output requirements:}
\begin{itemize}[leftmargin=*, itemsep=2pt, topsep=2pt]
  \item Output exactly one strict JSON array and nothing else.
  \item The array length must equal the number of analytical sentence records.
  \item Each item must be an object with:
  \begin{itemize}[leftmargin=1.5em, itemsep=1pt, topsep=1pt]
    \item \code{source_text}
    \item \code{propositions}
  \end{itemize}
  \item \code{propositions} must be a non-empty JSON array of strings.
\end{itemize}

\textbf{Example output.}
\begin{PromptCode}
[
  {
    "source_text": "This is a very practical decision",
    "propositions": ["This is a very practical decision"]
  }
]
\end{PromptCode}
\end{promptbox}
\begin{promptbox}{Prompt C.8: Analytical edge extraction}
\textbf{System Prompt.}
\begin{PromptCode}

\end{PromptCode}

\textbf{User Prompt.}
You analyze an analytical proposition batch and output only the sparse directed edges whose true anchor is the current source proposition.

\textbf{Relation type definitions.}
\begin{PromptCode}
\pvar^relation_standard~
\end{PromptCode}

\textbf{Rules:}
\begin{itemize}[leftmargin=*, itemsep=2pt, topsep=2pt]
  \item Use the full batch for context, but only output edges whose \code{source_proposition_id} equals the requested source proposition.
  \item Consider only forward edges to later propositions in the provided batch order.
  \item Treat all propositions uniformly; do not use sentence-boundary reasoning.
  \item Keep only clear, necessary, information-bearing edges.
  \item Do not output redundant shortcuts when a better downstream anchor exists.
  \item Do not output edges based only on same topic/entity mention or loose continuity.
  \item Output at most one relation per source-target pair.
  \item If no edge should be kept, output \code{[]}.
\end{itemize}

\textbf{Input Template.}
\begin{PromptCode}
Current source proposition id:
\pvar^source_proposition_id~

Proposition batch:
\pvar^proposition_batch_json~
\end{PromptCode}

\textbf{Output format.}
\begin{PromptCode}
[
  {
    "source_proposition_id": "ap_0000",
    "target_proposition_id": "ap_0003",
    "relation": "CAUSE",
    "explanation": "One short plain-text sentence only."
  }
]
\end{PromptCode}
\end{promptbox}
\begin{promptbox}{Prompt C.9: Analytical edge audit}
\textbf{System Prompt.}
\begin{PromptCode}

\end{PromptCode}

\textbf{User Prompt.}
You audit one candidate edge in an analytical proposition DAG.

\textbf{Relation type definitions.}
\begin{PromptCode}
\pvar^relation_standard~
\end{PromptCode}

Keep the candidate edge only if BOTH are clearly true:
\begin{enumerate}[leftmargin=*, itemsep=2pt, topsep=2pt]
  \item Correctness: the direction and relation are directly supported by the propositions.
  \item Necessity: the edge is necessary in the final sparse DAG, not a redundant shortcut or weak thematic link.
\end{enumerate}

\textbf{Guidance:}
\begin{itemize}[leftmargin=*, itemsep=2pt, topsep=2pt]
  \item Treat all proposition pairs uniformly; do not rely on sentence-boundary reasoning.
  \item Delete edges that are only temporal adjacency, same-topic mention, or narrative continuation.
  \item Delete broader shortcuts when a better downstream anchor already captures the relation.
  \item If not clearly convinced, delete.
\end{itemize}

\textbf{Allowed reason codes:} ... ...

\textbf{Input Template.}
\begin{PromptCode}
Full proposition list:
\pvar^all_propositions_json~

Draft edge list:
\pvar^draft_edges_json~

Candidate edge:
\pvar^candidate_edge_json~
\end{PromptCode}

\textbf{Output exactly one JSON object and nothing else.}
\begin{PromptCode}
{
  "keep": true,
  "is_correct": true,
  "is_necessary": true,
  "reason_code": "KEEP",
  "explanation": "One short plain-text sentence only."
}
\end{PromptCode}
\end{promptbox}
\begin{promptbox}{Prompt C.10: Analytical node and edge truth judgment}
\textbf{System Prompt.}
\begin{PromptCode}

\end{PromptCode}

\textbf{User Prompt.}
You judge whether each analytical proposition and each final DAG edge is true/supported in the commentary context.

\textbf{Important:}
\begin{itemize}[leftmargin=*, itemsep=2pt, topsep=2pt]
  \item Judge node/proposition truth and edge correctness, not extraction faithfulness.
  \item For an edge, use the given source node, target node, and relation together.
  \item Use the reference-event list as the primary evidence.
  \item Use the commentary only as necessary context for reference resolution.
\end{itemize}

\textbf{Mark a node true when:}
\begin{itemize}[leftmargin=*, itemsep=2pt, topsep=2pt]
  \item the proposition is supported by the reference-event list;
  \item the commentary context helps resolve pronouns or omitted referents without changing the claim;
  \item the proposition does not add unsupported event facts beyond the reference-event list.
\end{itemize}

\textbf{Mark a node false when:}
\begin{itemize}[leftmargin=*, itemsep=2pt, topsep=2pt]
  \item the proposition contradicts the reference-event list;
  \item the proposition depends on event facts that are not supported by the reference-event list;
  \item the commentary context does not supply enough support to make the proposition true.
\end{itemize}

\textbf{Mark an edge true only when:}
\begin{itemize}[leftmargin=*, itemsep=2pt, topsep=2pt]
  \item both endpoint propositions are supported;
  \item the directed relation and relation type are supported by the two propositions plus reference-event context.
\end{itemize}

\textbf{Mark an edge false when:}
\begin{itemize}[leftmargin=*, itemsep=2pt, topsep=2pt]
  \item either endpoint proposition is unsupported;
  \item the direction or relation type is unsupported;
  \item the link is only loose thematic, temporal, or narrative continuity.
\end{itemize}

\textbf{Input Template.}
\begin{PromptCode}
Commentary:
\pvar^commentary~

Candidate propositions:
\pvar^candidate_propositions_json~

Candidate final DAG edges:
\pvar^candidate_edges_json~

Reference-event list:
\pvar^reference_events_json~
\end{PromptCode}

\textbf{Output requirements:}
\begin{itemize}[leftmargin=*, itemsep=2pt, topsep=2pt]
  \item Output exactly one strict JSON object and nothing else.
  \item \code{node_labels} length must equal the number of candidate propositions.
  \item \code{edge_labels} length must equal the number of candidate final DAG edges.
\end{itemize}

\textbf{Example output.}
\begin{PromptCode}
{"node_labels": [true, false, true], "edge_labels": [false, true]}
\end{PromptCode}
\end{promptbox}

\subsubsection{Prompt for Background Evaluation}
\label{rubric2}

\begin{promptbox}{Prompt C.11: Background claim extraction (football)}
\textbf{System Prompt.}
You are a football background-claim extraction expert.

Extract medium-granularity background claims from football background commentary.

\textbf{Rules:}
\begin{itemize}[leftmargin=*, itemsep=2pt, topsep=2pt]
  \item Extract only claims supported by the source sentence.
  \item Use context only for disambiguation, entity resolution, and time-scope completion.
  \item Do not invent unsupported claims.
  \item Return self-contained, independently judgeable background claims.
  \item If no valid background claim exists, return no claim for that sentence.
\end{itemize}

\textbf{Allowed \code{claim_type} values:}
\begin{itemize}[leftmargin=*, itemsep=1pt, topsep=1pt]
  \item \code{identity_profile}
  \item \code{affiliation_history}
  \item \code{statistical_claim}
  \item \code{form_assessment}
  \item \code{match_metadata}
  \item \code{event_significance}
  \item \code{other_background}
\end{itemize}

\textbf{Allowed \code{focus_key} values include:}
... ...

\textbf{Allowed \code{time_scope} values:}
... ...

Each entity must contain \code{mention}, \code{resolved_name}, \code{entity_type}, and \code{resolution_confidence}. \code{resolution_confidence} should be \code{high} / \code{medium} / \code{low}.

Each claim object must include:
\begin{itemize}[leftmargin=*, itemsep=1pt, topsep=1pt]
  \item \code{claim_id}
  \item \code{source_sentence_id}
  \item \code{source_sentence}
  \item \code{claim_text}
  \item \code{claim_type}
  \item \code{focus_key}
  \item \code{entities}
  \item \code{time_scope}
\end{itemize}

Add \code{claimed_value} when the claim explicitly contains a numeric/statistical target. Add \code{qualifiers} when the claim requires extra qualifiers to be judgeable.

In batch mode, \code{source_sentence_id} must be the zero-based index in All Background Sentences. In per\_sentence mode, \code{source_sentence_id} must equal the Target Background Sentence ID. \code{source_sentence} must exactly copy the corresponding sentence text from All Background Sentences. Do not output legacy keys such as \code{claim/type} instead of \code{claim_text/claim_type}. Use \code{[]} when no valid background claim exists.

In batch mode, \code{source_sentence_id} must be the zero-based index in All Background Sentences. In per\_sentence mode, \code{source_sentence_id} must equal the Target Background Sentence ID. \code{source_sentence} must exactly copy the corresponding sentence text from All Background Sentences. Do not output legacy keys such as \code{claim/type} instead of \code{claim_text/claim_type}. Use \code{[]} when no valid background claim exists.

Return a strict JSON array only.

\textbf{User Prompt Template.}
\begin{PromptCode}
Extraction Mode: \pvar^extraction_mode~

Sport Type: \pvar^sport_type~

Target Background Sentence (ID=\pvar^source_sentence_id~):
\pvar^target_background_sentence~

Neighbor Background Sentences:
\pvar^neighbor_background_sentences~

All Background Sentences:
\pvar^all_background_sentences~

Descriptive Clauses:
\pvar^descriptive_clauses~

Analytical Clauses:
\pvar^analytical_clauses~

Parsed Match Context:
\pvar^parsed_match_context~

Extract zero or more background claims.
Every claim object must use the required schema from the system prompt.
In batch mode, use the zero-based index from All Background Sentences as source_sentence_id.
source_sentence must exactly copy that indexed sentence.
Return a strict JSON array only.
\end{PromptCode}
\end{promptbox}
\begin{promptbox}{Prompt C.12: Background factuality judgment}
\textbf{System Prompt.}
You are a factuality judge for background commentary claims.

Your task is to judge a single background claim as true or false.

\textbf{Core rules:}
\begin{itemize}[leftmargin=*, itemsep=2pt, topsep=2pt]
  \item The local background KB evidence package is the primary evidence source.
  \item You may use common knowledge only as a supplement, never as a replacement for stronger KB evidence.
  \item Every claim must receive a binary label: true or false.
  \item Prefer false when the available evidence does not justify a true judgement.
  \item Always return \code{evidence_ids} from the KB evidence items you actually used.
\end{itemize}

\textbf{Return strict JSON only with:}
\begin{PromptCode}
{
  "label": "true" or "false",
  "reasoning": "...",
  "evidence_ids": [...],
  "kb_priority_respected": true or false
}
\end{PromptCode}

\textbf{User Prompt Template.}
\begin{PromptCode}
Claim:
\pvar^claim~

Evidence Package:
\pvar^evidence_package~

Judge the claim as true or false.
The local KB evidence package must be treated as primary evidence.
Return strict JSON only.
\end{PromptCode}
\end{promptbox}
\begin{promptbox}{Prompt C.13: Background relevance judgment}
\textbf{System Prompt.}
You are a relevance matcher for background commentary claims.

Your task is to judge whether a background claim is relevant to the current clip, where the current clip is represented by descriptive clauses and analytical clauses.

\textbf{Rules:}
\begin{itemize}[leftmargin=*, itemsep=2pt, topsep=2pt]
  \item A claim is relevant if it meaningfully supports, explains, or contextualizes at least one descriptive or analytical clause.
  \item A claim is irrelevant if it does not meaningfully support either clause set.
  \item A single claim may match multiple descriptive clauses and multiple analytical clauses.
  \item Return exact clause ids you matched.
  \item If there is no meaningful match, return empty id lists and \code{label=irrelevant}.
\end{itemize}

\textbf{Return strict JSON only with:}
\begin{PromptCode}
{
  "label": "relevant" or "irrelevant",
  "matched_descriptive_sentence_ids": [...],
  "matched_analytical_sentence_ids": [...],
  "reasoning": "..."
}
\end{PromptCode}

\textbf{User Prompt Template.}
\begin{PromptCode}
Claim:
\pvar^claim~

Descriptive Clauses:
\pvar^descriptive_clauses~

Analytical Clauses:
\pvar^analytical_clauses~

Local Background Context:
\pvar^local_background_context~

Judge whether this background claim is relevant.
Return strict JSON only.
\end{PromptCode}
\end{promptbox}

\end{document}